\documentclass{article}

\PassOptionsToPackage{numbers, compress}{natbib}

\usepackage[preprint]{neurips_2023}

\usepackage[utf8]{inputenc} 
\usepackage[T1]{fontenc}    
\usepackage{hyperref}       
\usepackage{url}            
\usepackage{booktabs}       
\usepackage{amsfonts}       
\usepackage{nicefrac}       
\usepackage{microtype}      
\usepackage{xcolor}         

\usepackage{multirow}

\usepackage{graphicx}
\usepackage{amsmath}
\usepackage{amssymb}
\usepackage{booktabs}
\usepackage{xspace}
\usepackage{mathtools}
\usepackage{parskip}
\usepackage{enumitem}
\usepackage{algorithm}
\usepackage{listings}
\usepackage{bbm}
\usepackage{comment}

\usepackage[capitalize]{cleveref}
\crefname{section}{Sec.}{Secs.}
\Crefname{section}{Section}{Sections}
\Crefname{table}{Table}{Tables}
\crefname{table}{Tab.}{Tabs.}

\def\cvprPaperID{0009} 
\def\confName{CVPR}
\def\confYear{2023}

\begin{document}

\title{Peekaboo: Text to Image Diffusion Models are Zero-Shot Segmentors}

\def\modelname{Peekaboo\xspace}
\def\alphapenalty{Gravity\xspace}

\newcommand{\defeq}{\coloneqq}
\newcommand{\grad}{\nabla}
\newcommand{\E}{\mathbb{E}}
\newcommand{\Ea}[1]{\E\left[#1\right]}
\newcommand{\Eb}[2]{\E_{#1}\!\left[#2\right]}
\newcommand{\Vara}[1]{\Var\left[#1\right]}
\newcommand{\Varb}[2]{\Var_{#1}\left[#2\right]}
\newcommand{\kl}[2]{D_{\mathrm{KL}}\!\left(#1 ~ \| ~ #2\right)}
\newcommand{\bA}{\mathbf{A}}
\newcommand{\bI}{\mathbf{I}}
\newcommand{\bJ}{\mathbf{J}}
\newcommand{\bH}{\mathbf{H}}
\newcommand{\bL}{\mathbf{L}}
\newcommand{\bM}{\mathbf{M}}
\newcommand{\bP}{\mathbf{P}}
\newcommand{\bQ}{\mathbf{Q}}
\newcommand{\bR}{\mathbf{R}}
\newcommand{\bW}{\mathbf{W}}
\newcommand{\bzero}{\mathbf{0}}
\newcommand{\bone}{\mathbf{1}}
\newcommand{\bb}{\mathbf{b}}
\newcommand{\bc}{\mathbf{c}}
\newcommand{\bd}{\mathbf{d}}
\newcommand{\be}{\mathbf{e}}
\newcommand{\bh}{\mathbf{h}}
\newcommand{\bu}{\mathbf{u}}
\newcommand{\bv}{\mathbf{v}}
\newcommand{\bw}{\mathbf{w}}
\newcommand{\bx}{\mathbf{x}}
\newcommand{\by}{\mathbf{y}}
\newcommand{\bz}{\mathbf{z}}
\newcommand{\bxh}{\hat{\mathbf{x}}}
\newcommand{\btheta}{{\boldsymbol{\theta}}}
\newcommand{\bphi}{{\boldsymbol{\phi}}}
\newcommand{\bepsilon}{{\boldsymbol{\epsilon}}}
\newcommand{\bmu}{{\boldsymbol{\mu}}}
\newcommand{\bnu}{{\boldsymbol{\nu}}}
\newcommand{\bSigma}{{\boldsymbol{\Sigma}}}
\newcommand{\lsd}{\mathcal{L}_{s}}
\newcommand{\gravity}{\mathcal{L}_{\alpha}}
\newcommand{\lpeekaboo}{\mathcal{L}_{p}}

\newcommand{\alphamask}{\boldsymbol{\alpha}}
\author{%
  Ryan Burgert \quad
  Kanchana Ranasinghe \quad
  Xiang Li \quad
  Michael S. Ryoo
  \vspace{0.5em} \\
  Stony Brook University \quad 
  \vspace{0.2em} \\
  \small{\texttt{rburgert@cs.stonybrook.edu}}
}


\maketitle

\begin{abstract}
Recently, text-to-image diffusion models have shown remarkable capabilities in creating realistic images from natural language prompts. However, few works have explored using these models for semantic localization or grounding. In this work, we explore how an off-the-shelf text-to-image diffusion model, trained without exposure to localization information, can ground various semantic phrases without segmentation-specific re-training. We introduce an inference time optimization process capable of generating segmentation masks conditioned on natural language prompts. Our proposal, \modelname, is a first-of-its-kind zero-shot, open-vocabulary, unsupervised semantic grounding technique leveraging diffusion models without  any training. We evaluate \modelname on the Pascal VOC dataset for unsupervised semantic segmentation and the RefCOCO dataset for referring segmentation, showing results competitive with promising results. We also demonstrate how \modelname can be used to generate images with transparency, even though the underlying diffusion model was only trained on RGB images - which to our knowledge we are the first to attempt. 
Please see our project page, including our code: \url{https://ryanndagreat.github.io/peekaboo}
\end{abstract}

\section{Introduction}

\label{sec:intro}

Image segmentation, a key computer vision task, involves dividing an image into meaningful spatial regions. Semantic segmentation assigns pre-defined labels \cite{coco}, while referring segmentation allows any natural language prompt \cite{hu2016segmentation}. Both tasks are essential for real-world applications \cite{sun2020scalability}.
Recent progress in semantic segmentation \cite{chen2017deeplab,chen2018searching} relies on expensive manual annotations, but weak supervision approaches \cite{Ghiasi2021OpenVocabularyIS,li2022language} leveraging contrastive image language pre-training models \cite{clip,jia2021scaling} have emerged. Referring segmentation has adopted language-specific components \cite{Wang2022CRISCR} based on \cite{clip}. Unsupervised semantic segmentation approaches \cite{Xu2022GroupViTSS,Ranasinghe2022PerceptualGI} struggle with complex language prompts, particularly in referring segmentation tasks (\cref{tbl:refcoco}).

Despite contrastive image language pre-training models \cite{clip} serving as a foundation for segmentation tasks \cite{Ghiasi2021OpenVocabularyIS, Wang2022CRISCR}, diffusion models \cite{ddpm, pmlr-v37-sohl-dickstein15, scoresde} have not been utilized for segmentation with the exception of \cite{ODISE}, which requires expensive compute to train. We ask if pre-trained diffusion models can associate natural language with relevant spatial regions of an image. Our proposed \modelname, based on a pre-trained image-language stable diffusion model \cite{Rombach2022HighResolutionIS}, achieves unsupervised semantic and referring segmentation without having to training any model.
\modelname employs an inference time optimization that iteratively updates an alpha mask, converging to optimal segmentation for a given image and language caption. We propose a novel alpha compositing-based loss for improved learning of the alpha mask.
Our contributions are:
\begin{enumerate}[topsep=-0.5ex,itemsep=-0.5ex,partopsep=0ex,parsep=1ex]
\item Novel mechanism for unsupervised segmentation applicable to semantic and referring segmentation tasks
\item Demonstrating pixel-level localization information within pre-trained text-to-image diffusion models
\item A mechanism for using Stable Diffusion as an off-the-shelf foundation model for  segmentation
\item End-to-end text-to-image generation of RGBA images with transparency, which to our knowledge has not been previously attempted.
\end{enumerate}

We evaluate our approach on both
RefCOCO \cite{Kazemzadeh2014ReferItGameRT}
 and modified Pascal VOC \cite{everingham2010pascal} (Pascal VOC-C).

\begin{figure}[t]
\centering

\includegraphics[width=.5\linewidth]{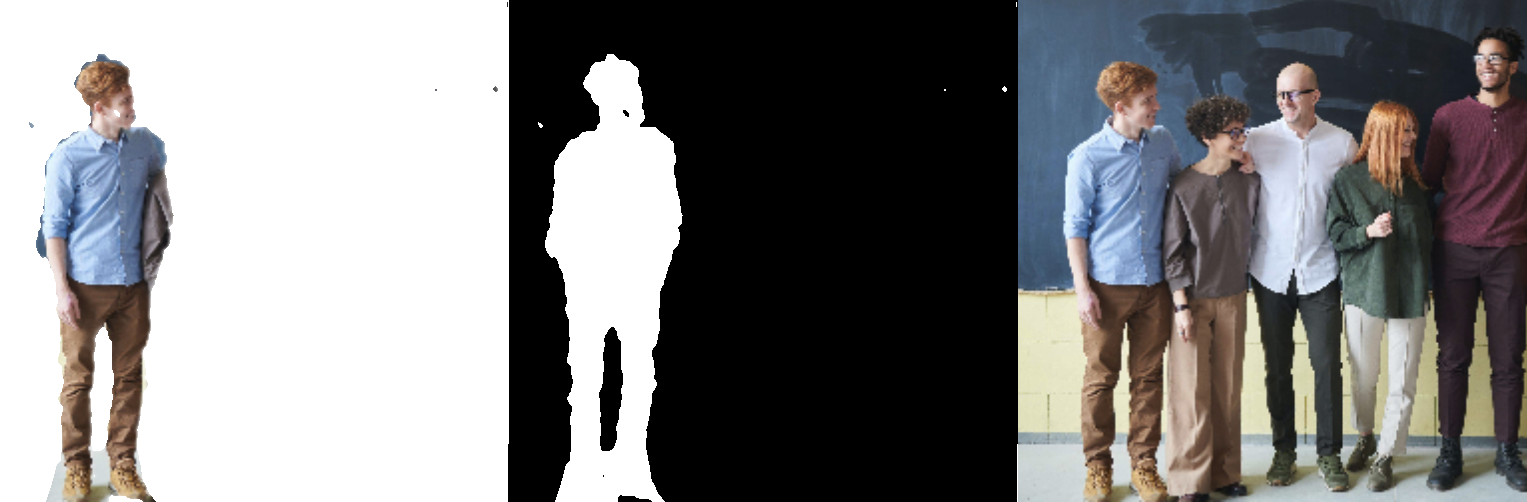}\\
\texttt{man with blonde hair in blue shirt and brown pants}

\includegraphics[width=.5\linewidth]{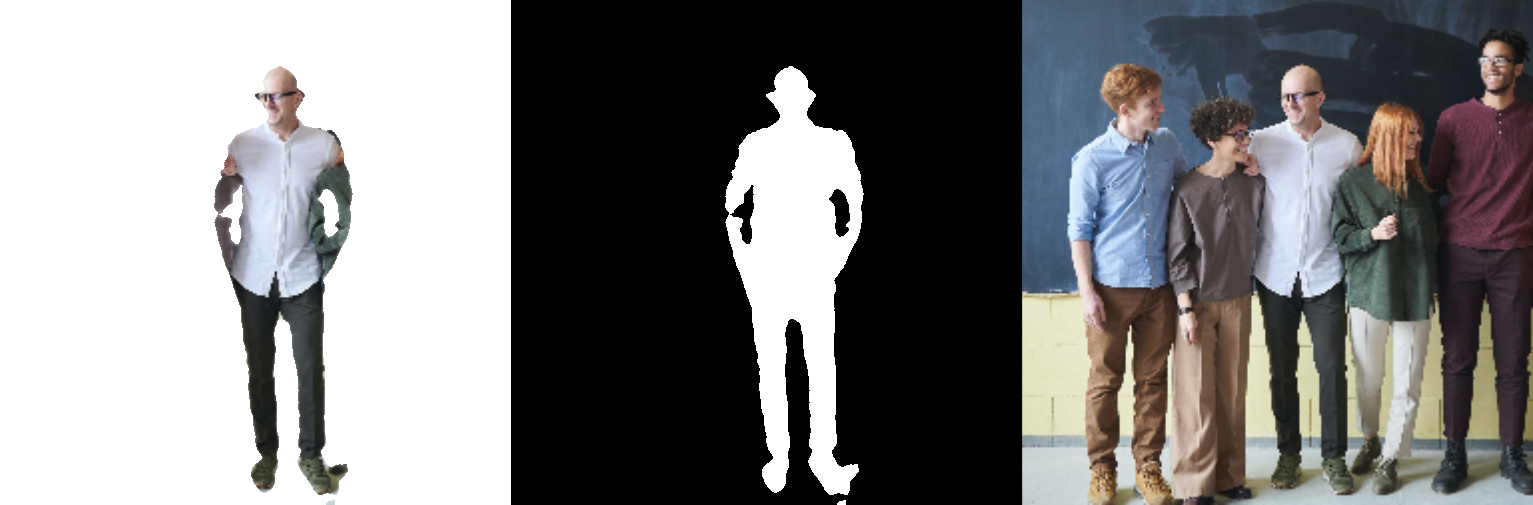}\\
\texttt{bald man wearing glasses with white shirt and black pants}

\caption{\textbf{Zero-Shot Segmentation with phrases}:  We highlight the ability of \modelname to ground complex language prompts onto an image with no segmentation specific training. An off-the-shelf diffusion model is used with only an inference time optimization technique to generate these segmentations. If you look closely, you'll notice the bald man's arms are segmented - but are not visible in the photo! Peekaboo has fairly strong shape priors.}
\label{fig:intro}
\end{figure}

\begin{figure*}[t]
\centering
\scalebox{0.7}{
	\begin{minipage}{0.3\linewidth}
		\definecolor{amber}{rgb}{1.0, 0.75, 0.0}
		\includegraphics[width=\linewidth]{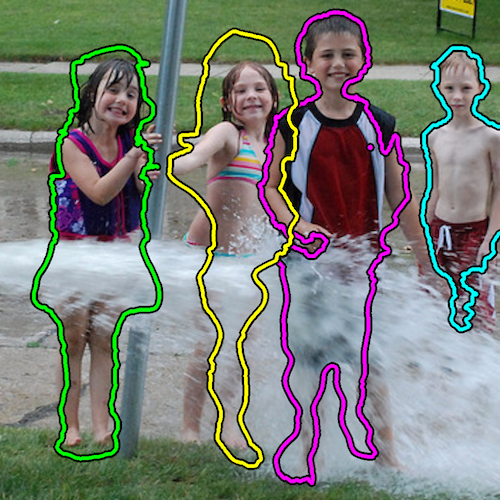}
		\small{
		\textbf{\textcolor{green}{\texttt{girl left in purple}}} \\
		\textbf{	\textcolor{magenta}{\texttt{boy in red vest}}} \\
		\textbf{	\textcolor{amber}{\texttt{small girl with striped bikini}}} \\
		\textbf{	\textcolor{cyan}{\texttt{small shirtless boy on the right}}}}
	\end{minipage}
	\begin{minipage}{0.69\linewidth}
		\includegraphics[width=0.32\linewidth]{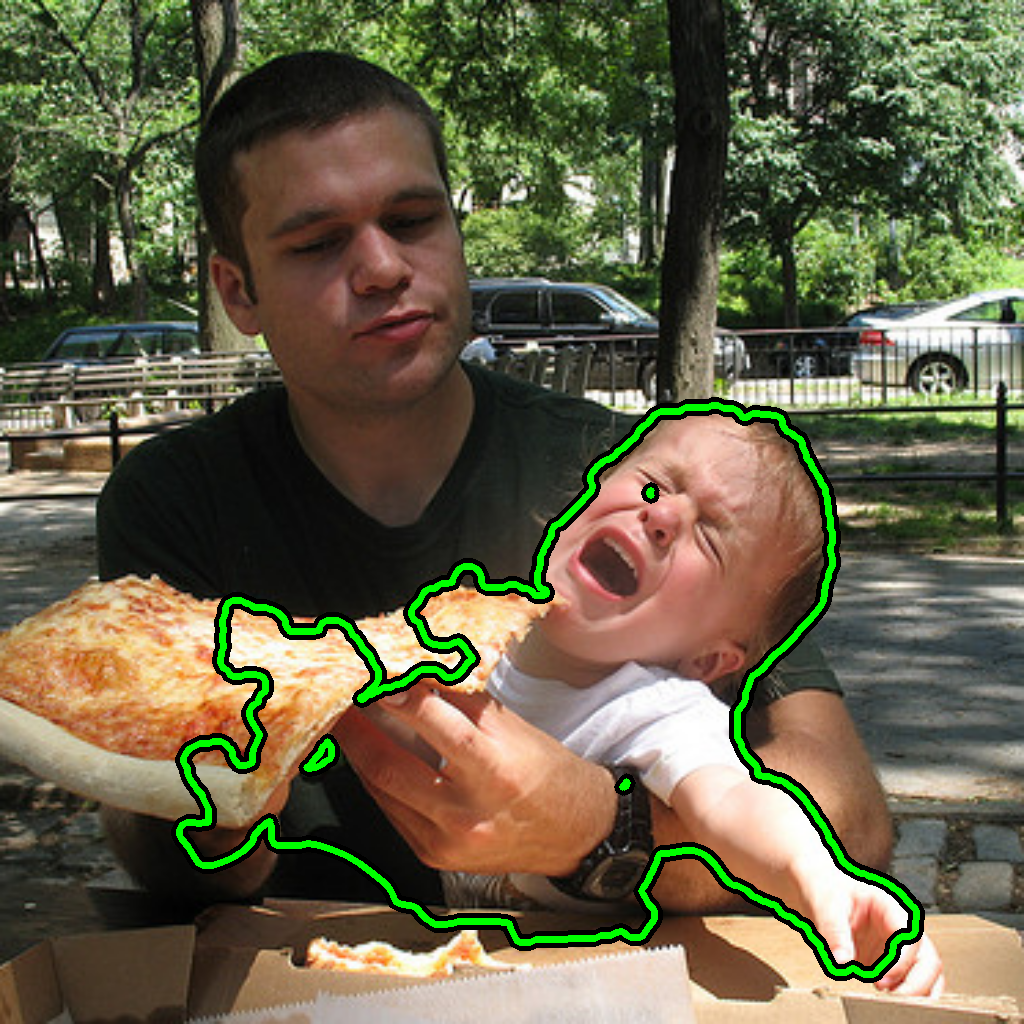}
		\includegraphics[width=0.32\linewidth]{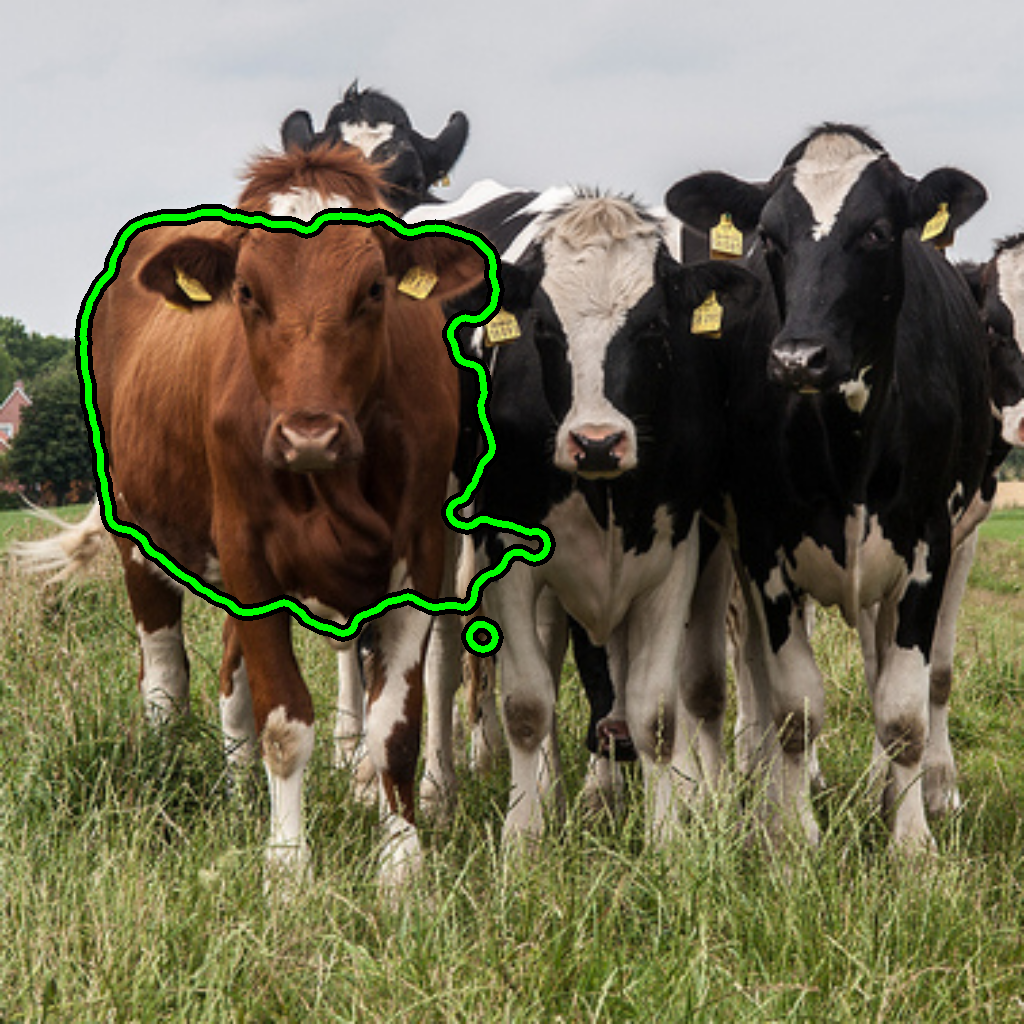}
		\includegraphics[width=0.32\linewidth]{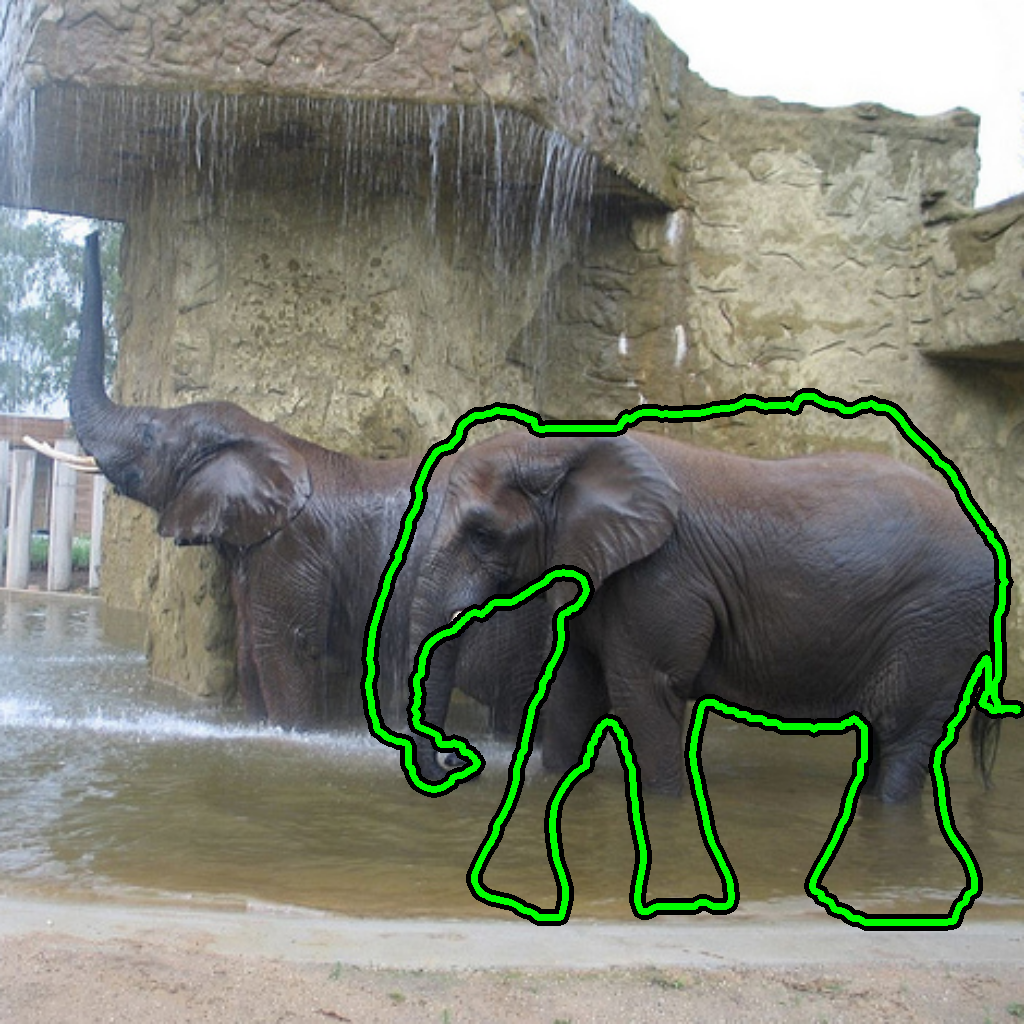} \\
		\includegraphics[width=0.32\linewidth]{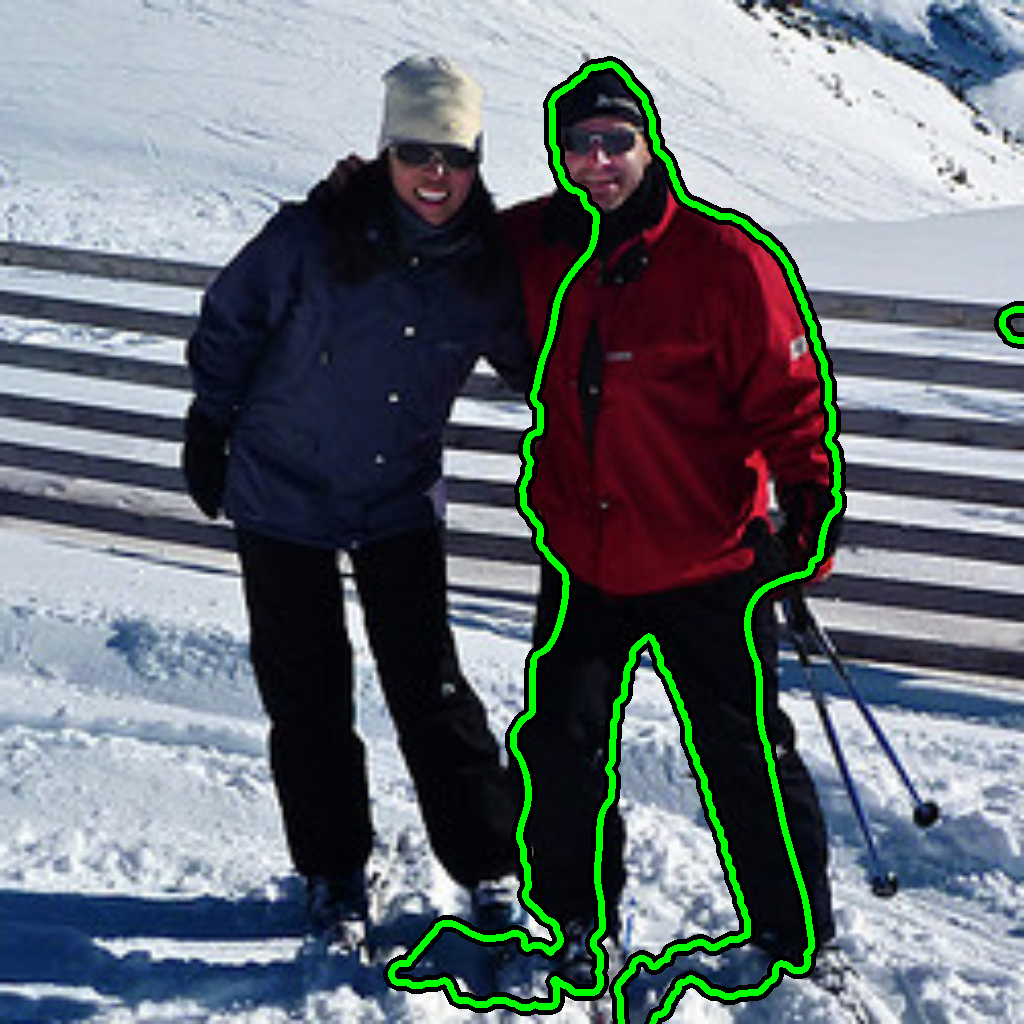}
		\includegraphics[width=0.32\linewidth]{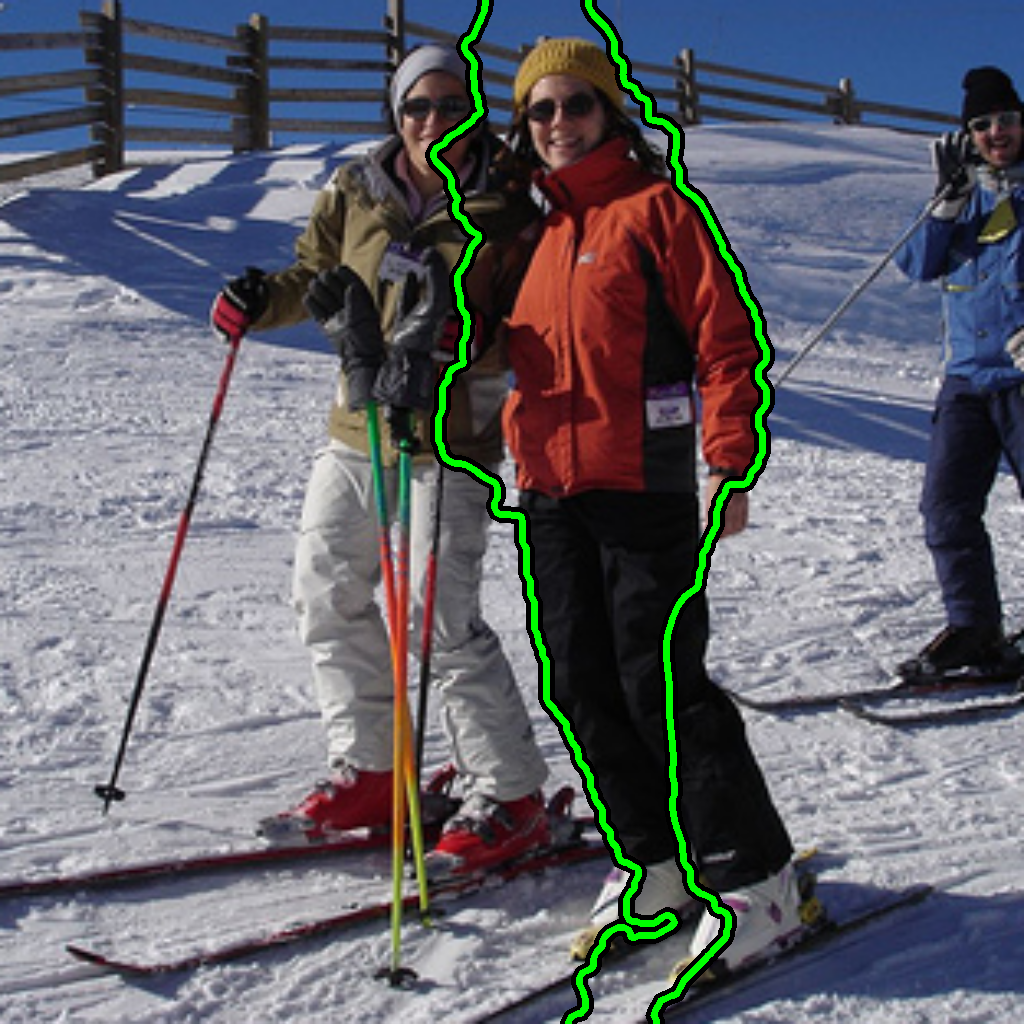}
		\includegraphics[width=0.32\linewidth]{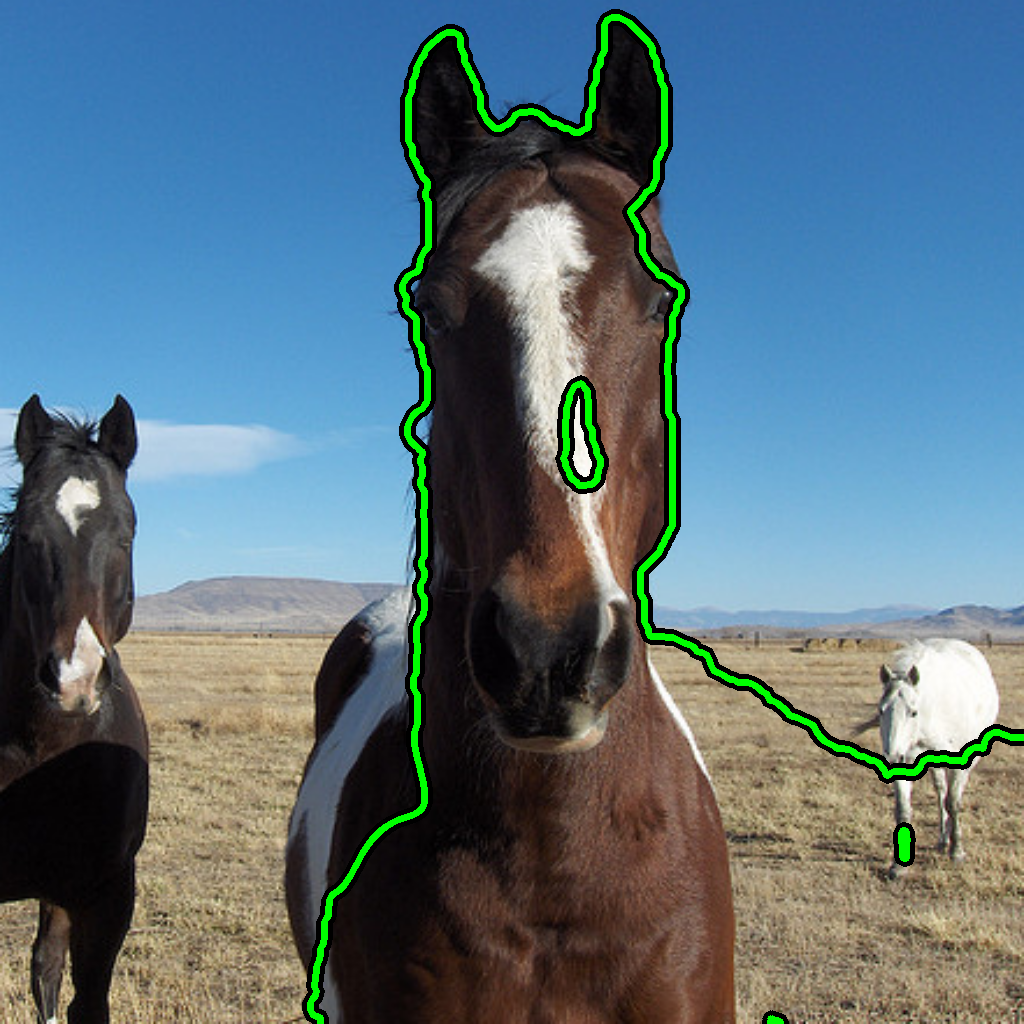}
	\end{minipage}
}
	\vspace{-0.5em}
	\caption{\textbf{ \modelname Samples}:
	(left) we apply referring segmentation for each object in an example image from RefCOCO using custom text prompts listed below the image. (right) We illustrate 6 more examples from RefCOCO using their existing captions, listed left to right for top and bottom respectively: "baby", "brown cow", "elephant on right", "guy in red", "red jacket", "the front horse".}
	\label{fig:vis}
	\vspace{-0.5em}
\end{figure*}

\section{Related Work}
\label{sec:related}
\textbf{Vision-Language Models:} Vision-language models have advanced rapidly, enabling zero-shot image recognition \cite{frome2013devise,socher2013zero} and language generation from visual inputs \cite{karpathy2015deep,vinyals2015show,kiros2014unifying,mao2014explain}. Recent contrastive language-image pre-training models \cite{clip,jia2021scaling} showcase open-vocabulary and zero-shot capabilities, with extensions for a wide range of tasks \cite{pham2021combined,yu2022coca, desai2021virtex, yao2022filip,cui2022democratizing,Zeng2022SocraticMC, yuan2021florence, kamath2021mdetr}. Grounding language to images \cite{Gu2022OpenvocabularyOD, Ghiasi2021OpenVocabularyIS} and unsupervised segmentation \cite{Xu2022GroupViTSS, Ranasinghe2022PerceptualGI} have also been explored. Our proposed \modelname leverages an off-the-shelf diffusion model without segmentation-specific re-training and handles sophisticated compound phrases.

\textbf{Diffusion Models:} Diffusion Probabilistic Models \cite{pmlr-v37-sohl-dickstein15} have been adopted for language-vision generative tasks \cite{Nichol2022GLIDETP,dalle,dalle2,imagen,palette,parti,sr3} and extended to various applications \cite{wavegrad,videodiffusion,diffwave, Poole2022DreamFusionTU}. Recent works \cite{Poole2022DreamFusionTU, graikos2022diffusion} sample diffusion models through optimization. Our \modelname utilizes efficient latent diffusion models and is the first zero-shot method for cross-modal discriminative tasks such as segmentation.

\textbf{Score Distillation Loss:} SDL was introduced in DreamFusion \cite{Poole2022DreamFusionTU} and applied to NeRF \cite{mildenhall2020nerf} to create 3D models. Our work applies it to alpha masks for image segmentation, yielding better results than previous CLIP-based techniques \cite{crowson2022vqganclip,jain2021dreamfields,khalid2022clipmesh}. Like Peekaboo, SDL is also used with Stable Diffusion in \cite{burgert2023diffusion_illusions,lin2022magic3d,wordasimage}.

\textbf{Unsupervised Segmentation:} Unsupervised segmentation \cite{malik2001visual} has evolved from early spatially-local affinity methods \cite{comaniciu1997robust,shi2000normalized,ren2003learning} to deep learning-based self-supervised approaches \cite{caron2021emerging, hamilton2022unsupervised, Cho2021PiCIEUS, VanGansbeke2021UnsupervisedSS, Ji2019InvariantIC}. LSeg \cite{li2022language} is a semi-supervised segmentation algorithm because it's trained with ground truth segmentation masks, but attempts to generalize the dataset labels to language using CLIP embeddings. Our \modelname also enables grouping aligned to natural language, is open-vocabulary.

\textbf{Referring Segmentation:} Referring segmentation \cite{hu2016segmentation,yu2018mattnet,ye2019cross} involves vision and language modalities. Early approaches \cite{hu2016segmentation,liu2017recurrent,li2018referring,margffoy2018dynamic} fuse features, while recent works use attention mechanisms \cite{chen2019referring,shi2018key,ye2019cross,huang2020referring,huilinguistic} and cross-modal pre-training \cite{Wang2022CRISCR}.
Large supervised segmentation models \cite{SAM,SEEM} have demonstrated high performance, but they require annotated segmentation datasets. Concurrently, an unsupervised referring segmentation method \cite{ODISE} has been developed using the same Stable Diffusion model as us, but it requires significant computational resources, including 5.3 days of training with 32 NVIDIA V100 GPUs. 
In contrast, \modelname is the first to perform unsupervised referring segmentation without necessitating any model training, effectively reducing the training time to 0 days on 0 GPUs.

\begin{figure*}[t]
\centering

\includegraphics[width=0.90\textwidth]{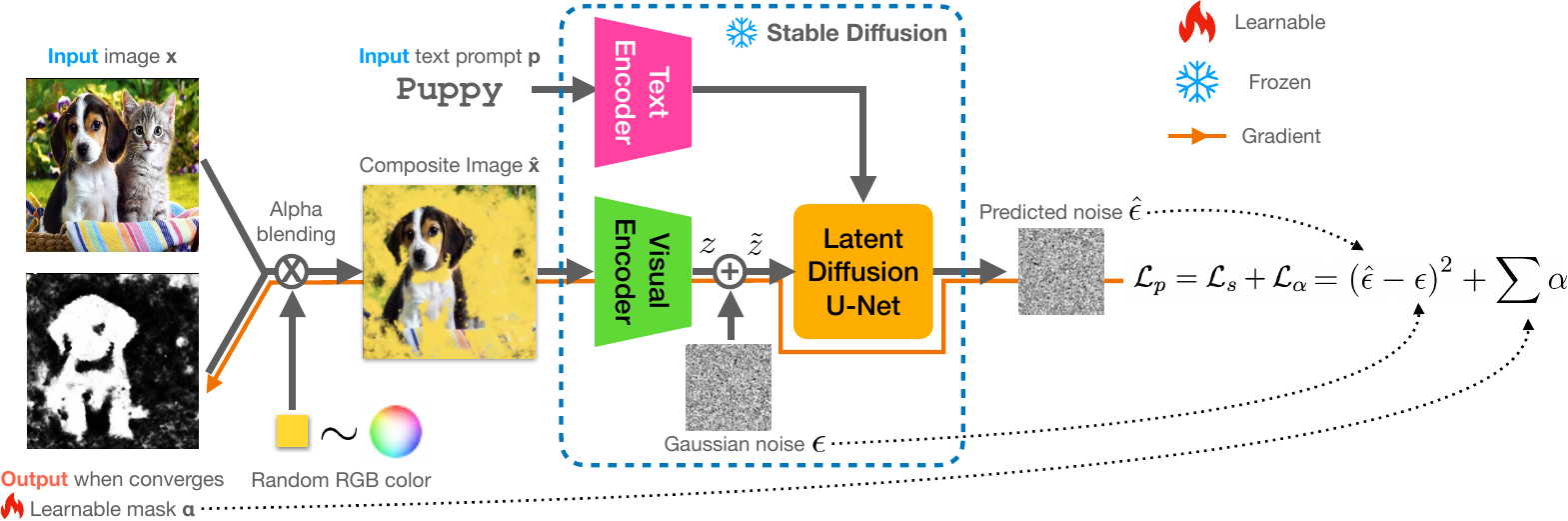}
\vspace{-0.5em}
\caption{\textbf{Overview of \modelname Architecture}: 
We illustrate how an input image and random background are alpha blended to generate a composite image. This image and its relevant text prompt are processed by our diffusion model based inference-time objective. Iterative gradient based optimization of the randomly initialized alpha mask converges to a segmentation optimal for the conditioning text prompt. Note that our diagram shows the alpha mask at an intermediate iteration: at the initial iteration it is entirely random Gaussian noise.   
}
\label{fig:arch}
\vspace{-1.0em}
\end{figure*}
\section{Proposed Method}

\begin{figure*}[t]
	\begin{minipage}{\linewidth}
		\includegraphics[width=0.96\linewidth]{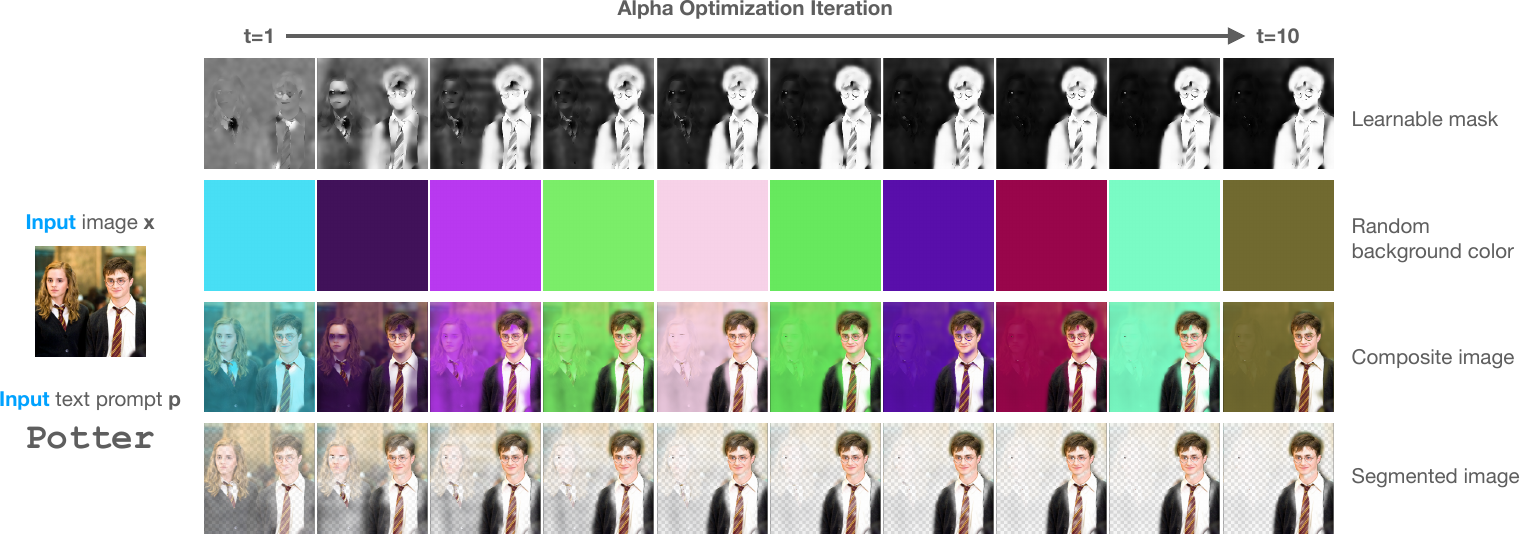}
	\end{minipage}
	\caption{\textbf{Inference Process}:
	Above we show a timelapse of the inference-time alpha mask optimization process
	}
	\label{fig:vis_more}
\end{figure*}
\label{sec:method}
In this section, we discuss our \emph{zero-training} approach for \emph{unsupervised zero-shot} segmentation, \modelname. We formulate segmentation as a foreground alpha mask optimization problem and leverage a text-to-image stable diffusion model pre-trained on large internet-scale data. The alpha mask is optimized with respect to image and text prompts.

\subsection{Background: Score Distillation Sampling} 
\label{subsubsec:sds}
First introduced in DreamFusion \cite{Poole2022DreamFusionTU}, Score Distillation Sampling (SDS) is a method that generates samples from a diffusion model by optimizing a loss function we call \textit{score distillation loss} (SDL). This allows us to optimize samples in any parameter space, as long as we can map back to images in a differentiable manner. We modify SDS to optimize learnable alpha masks and operate with latent diffusion.

\subsection{Overview}
\label{subsec:arch}
Consider an image of a puppy. Its defining region is the foreground (region containing the puppy): all distinctive characteristics of the puppy are retained even when the background is completely altered. \modelname leverages this idea to iteratively optimize a learnable alpha mask such that it converges to the optimal foreground segmentation. 
We use a randomly initialized alpha mask ($\alphamask$) to alpha-blend the puppy image ($\bx$) with different backgrounds ($\bb$), generating new composite images ($\hat{\bx}$) as described in \cref{eq:blend}:
\begin{equation}
\label{eq:blend}
    \hat{\bx} = \alphamask \bx + (1-\alphamask) \bb
\end{equation}
The composite image and related text prompt ($\mathbf{p}$) are jointly processed by a pre-trained text-to-image diffusion model. An SDL based objective building off its output is minimized by iteratively optimizing the alpha mask. Minimizing this inference-time objective makes each new composite image similar to the text prompt (e.g. \textit{puppy}) in latent space. We next discuss this inference-time objective in detail. 

\subsection{Inference-time Objective}
\label{subsec:loss}
\modelname's inference-time objective $\lpeekaboo$ has two components: {latent score distillation loss} $\lsd$ and {alpha regularization loss} $\gravity$. The total loss, $\lpeekaboo=\lsd+\gravity$, is called the {Peekaboo loss}.

\textbf{Latent Score Distillation Loss} or $\lsd$ can be conceptually interpreted as a measurement of cross-modal similarity between a composite image and a text prompt $\mathbf{p}$. 
The key intuition lies in how predicted noise from the diffusion model is minimal for samples better matching the conditional distribution.
\modelname utilizes Stable Diffusion that operates in a latent space. We adapt standard SDL to operate within this latent space, hence the term \emph{latent} score distillation loss.

The Stable Diffusion model jointly processes images and text. Its visual encoder first projects images to a latent space. This latent vector is then processed by the diffusion U-Net ($\mathcal{D}$) conditioned on text embedding (from text encoder $\mathcal{T}$) to produce noise outputs.   
To measure $\lsd$, we first degrade latent vector $z$ of composite image $\bx$ using forward diffusion, introducing Gaussian noise $\epsilon \sim \mathcal{N}$ to $z$, resulting in a noisy $\Tilde{z}$.
We then perform diffusion denoising conditioned on the text embedding with pre-trained $\mathcal{D}$. 
Our loss $\lsd$ is measured as the reconstruction error of noise $\epsilon$, given noisy $\Tilde{z}$ and text embedding $\mathcal{T}\left( \mathbf{p} \right)$ as in \cref{eq:lsdl}:
\begin{equation}
\label{eq:lsdl}
    \lsd = \mathrm{MSE}\left( \epsilon, \mathcal{D}\left(\Tilde{z}, \mathcal{T}\left( \mathbf{p} \right) \right) \right)
\end{equation}
where MSE refers to mean-squared loss. Pseudo-code describing $\lsd$ in detail is presented in \cref{alg:peekaboo}.

\begin{algorithm}[t]

\caption{Pytorch style pseudocode for Peekaboo}
\label{alg:peekaboo}

\definecolor{codeblue}{rgb}{0.25,0.5,0.5}
\definecolor{codeorange}{rgb}{1.0,0.5,0.3} 
\definecolor{codegreen}{rgb}{0.13,0.54,0.13}
\lstset{
  basicstyle=\fontsize{7.2pt}{7.2pt}\ttfamily\bfseries,
  commentstyle=\fontsize{7.2pt}{7.2pt}\color{codeblue},
  keywordstyle=\fontsize{7.2pt}{7.2pt}\color{codeorange},
  stringstyle=\color{codegreen},
  numbers=left,  
  numbersep=5pt, 
  numberstyle=\fontsize{6.2pt}{6.2pt}\color{codeblue},
  columns=fixed,
  xleftmargin=2em,
  framexleftmargin=1.5em
}

\begin{lstlisting}[language=python] 
import torch, stable_diffusion as sd

def segment_image_via_peekaboo(img, prompt):
  """Given an image and text prompt, return an alpha 
  mask that is close to 1 in regions where prompt 
  is relevant and 0 where the prompt is not."""
  
  alpha = LearnableAlphaMask(img) # nn.Module 
  optim = torch.optim.SGD(alpha.parameters())
  
  for _ in range(num_iterations):
      peekaboo_loss(img, prompt, alpha()).backward()
      optim.step() ; optim.zero_grad()
  return alpha()

def peekaboo_loss(img, prompt, alpha):
  """Core of our paper. Blends image with random 
  color and returns loss guiding alpha mask."""
  
  img_embedding = sd.vae.encoder(img)
  
  background = torch.random(3) # random RGB color
  composite_img = torch.lerp(background, img, alpha)
  
  loss = alpha_regularization_loss = alpha.sum()
  loss += score_distill_loss(img_embedding, prompt)
  
  return loss

def score_distill_loss(image_embedding, prompt):
  """Same loss proposed in DreamFusion"""
  timestep = random_int(0, _diffusion_step)
  noise = sd.get_noise(timestep)
  noised_embed = \
    sd.add_noise(image_embedding, noise, timestep)  
  with torch.no_grad():
      text_embed = sd.clip.embed(prompt)    
      predicted_noise = \ 
        sd.unet(noised_embed, text_embed, timestep)  
  return (torch.abs(noise - predicted_noise)).sum()

class LearnableAlphaMask(nn.Module):
  """This class parameterizes the alpha mask"""
  def __init__(self, image):
    _, H, W = image.shape
    self.alpha = nn.Parameter(torch.random(1, H, W))
    self.img = image
  def forward(self):
    alpha = bilateral_blur(self.alpha, self.img)
    return torch.sigmoid(alpha)
\end{lstlisting}
\end{algorithm}
\label{sec:alpha_reg}
\textbf{Alpha Regularization Loss} or $\gravity$ enforces a minimal alpha mask. Specifically, $\gravity = \sum_i \alphamask_i$, where $i$ indexes pixel location in $\alphamask$.
\label{subsec:aux_loss}
Assuming a target is in our image, a trivial way to satisfy cross-modal similarity ($\lsd$) is to include all pixels. That is, setting $\alphamask = \mathbbm{1}^{H\times W}$ (all ones array).
In this case, composite image $\hat{\bx}$ would satisfy the prompt since $\hat{\bx}$ is identical to input image $\bx$.
To combat this, we penalize high $\alphamask$ values with regularization, discouraging unnecessary pixels from being added. 

\subsection{Alpha Mask Parametrization}
We next discuss the parametrization of our learnable alpha mask. Our experiments indicate that representing an alpha mask as a simple learnable matrix (referred as \textit{raster} parametrization) yields sub-optimal results. To improve this, we apply a bilateral blur to that matrix and then clip it between 0 and 1 using a sigmoid function. 
This approach, defined \textit{raster bilateral} parametrization, allows us to align our generated segmentation masks with the image content at a pixel level, respecting boundaries between regions present in the image. 
As a result, we are able to achieve better segmentation. We also investigate alternative parametrizations in section \ref{sec:experiments}.

\subsubsection{Peekaboo's Bilateral Filter}
\label{sec:bilateralfiltersubsub}

In this subsection, we provide a detailed exploration of the bilateral filter, briefly introduced above.

A standard bilateral filter is a non-linear filter that smooths an image while preserving its edges. It does this by considering both the spatial distance and color differences between pixels, giving higher weight to pixels that are close in space and have similar colors.

We apply a modified version of this filter onto the learnable alpha mask. This filter operates on the alpha mask tensor and modifies it using the color and spatial information of the image to be segmented. This filter is applied at every step the alpha  optimization process, as opposed to during post-processing.

The intention behind using such a filter is to ensure that our generated segmentation masks adhere to the image content at a pixel level from early iterations, leading to improved segmentation in the end.

\label{sec:experiments}
In this section, we present results, both quantitative and qualitative for our downstream tasks of segmentation. We first go over peekaboo and our baseline algorithms, then talk about the specific experiments.

\subsection{Baselines}
	Following \cite{hu2016segmentation}, we build a baseline that predicts the entire image as the segmentation tagged \textit{Random (whole image)}. For our next baseline, we apply the weakly-supervised segmentation method GroupViT \cite{Xu2022GroupViTSS}. Note that GroupViT is intended for text conditioned segmentation and is trained on datasets similar to those used by Stable Diffusion (i.e. our off-the-shelf diffusion model). Our last baseline, LSeg \cite{li2018referring} is a semi-supervised segmentation alogorithm that has seen the VOC dataset during its training. 

\newcommand{\vart}[1]{\textit{``#1''}}
\subsection{Peekaboo Variants}
\begin{figure}[t]
\centering
	\includegraphics[width=0.85\linewidth]{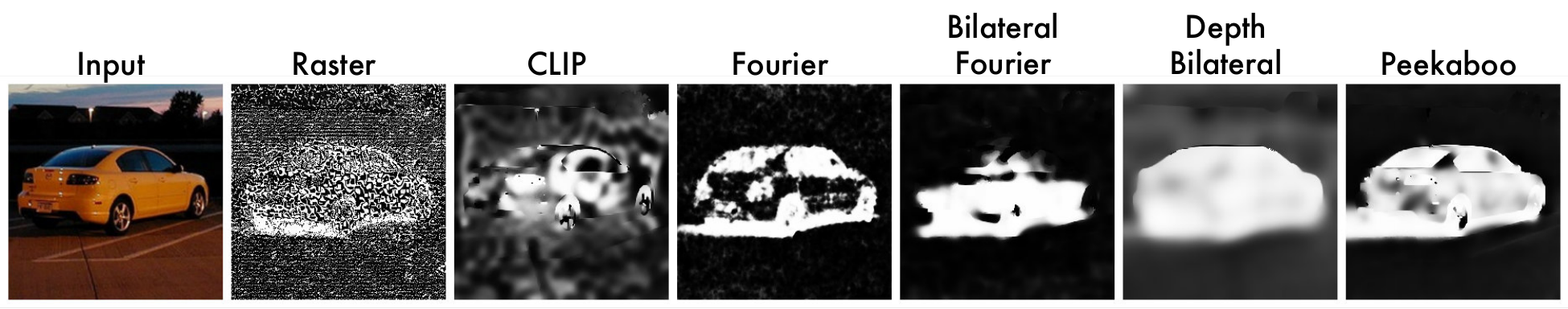}
	\caption{\textbf{Visual Variant Comparison}:
	This figure gives qualitative comparisons between Peekaboo variants on a single image from VOC-C with the prompt ``car''.
	}
	\label{fig:visualvariations}
\end{figure}

	We our experiments we showcase several different variants and ablations of Peekaboo.  \cref{fig:visualvariations} gives a qualitative comparison between these variants.
 
    \vart{RGB Bilateral} is the main, default Peekaboo algorithm described in Section \ref{sec:method}. For more details on these methods, please see the appendix.
	\vart{CLIP}: this variant substitutes score distillation loss for a CLIP-based \cite{clip} loss similar to \cite{jain2021dreamfields}. 
	All of the other variants are changes solely to the alpha mask parametrization.
	\vart{Raster}: this ablation skips the bilateral filter, and thus optimizes the alpha map pixel-wise. This yields worse performance than any other parametrization.
	\vart{Fourier}: this variant parametrizes the alpha mask with a fourier feature network \cite{ffn}, inspired by neural neural textures in \cite{Burgert2022}. This variant does not use a bilateral filter, but yields better results than a basic matrix parametrization. It tends to suffer from hallucinations more than other variants.
	\vart{Fourier Bilateral}: Just like the Fourier variant, except with the bilateral filter on top. This yields higher performance than the fourier feature network alone.
	\vart{Depth Bilateral}: Uses a depth map generated using the off-the-shelf monocular depth estimation model MIDAS \cite{midas} to guide the bilateral filter instead of differing RGB values. Intuitively, it means that pixels that are close in 3d space will have similar alpha values. This variant performs better than the main Peekaboo algorithm, and outperforms Clippy  \cite{Ranasinghe2022PerceptualGI} on both COCO and VOC-C.

\section{Experiments}

\subsection{Referring Segmentation}

We first present our quantitative evaluations for referring segmentation in \cref{tbl:refcoco}. 
These results presented in \cref{tbl:refcoco} showcase impressive performance of \modelname. 

The RefCOCO dataset is farily challenging, as the prompts are complex refer to a specific portion of the image. Some example prompts: ``giant metal creature with shiny red eyes'', ``bartender at center in gray shirt and blue jeans'', and ``suitcase behind the zebra bag''. 


GroupViT is a key comparison to our work, given how it is trained on similar noisy image-caption pair from internet scraped data. And on this dataset, Peekaboo outperforms GroupViT - a model that was trained specifically for segmentation on 16 NVIDIA V100 GPUs for two days. LSeg was trained for segmentation and outperforms Peekaboo, but we would like to reiterate - \modelname simply utilizes a diffusion model originally trained for text-to-image generation, and adopts it to segmentation with no re-training.



\subsection{Semantic Segmentation}

\begin{table}[t]
	\centering
	\begin{tabular}{c|c|c|c|c|c|c}
		\toprule
		\textbf{Type} & \textbf{Method} & \textbf{Prec@0.2} & \textbf{Prec@0.4} & \textbf{Prec@0.6} & \textbf{Prec@0.8} & \textbf{mIoU} \\ 
  \midrule
        \multirow{3}{*}{Baselines} 
        & Random & .141 & .022 & .003 & .000 & .102 \\
		& GroupVIT \cite{Xu2022GroupViTSS} & .212 & .075 & .020 & .002 & .112 \\
		& LSeg \cite{li2022language} & .512 & .212 & .051 & .008 & .235 \\
  \midrule
        \multirow{2}{*}{\begin{tabular}{c}\textbf{Peekaboo}\\Variants (ours)\end{tabular}} 
		& Depth Bilateral & .359 & .135 & .037 & .003 &  .204 \\
		& \textbf{RGB Bilateral} & .318 & .099 & .018 & .002 & .163 \\
        \bottomrule
	\end{tabular}
	\caption{\textbf{Referring Segmentation Evaluation.} We present results on the RefCOCO dataset. Numbers reported are precision and mIoU values for our method and baselines. Once more, unlike the others, \modelname is without any segmentation training.}
	\label{tbl:refcoco}
	\vspace{-0.5em}
\end{table}
\begin{table}[t]
	\centering
	\begin{tabular}{c|c|c|c|c|c|c}
		\toprule
		\textbf{Type} & \textbf{Method} & \textbf{Prec@0.2} & \textbf{Prec@0.4} & \textbf{Prec@0.6} & \textbf{Prec@0.8} & \textbf{mIoU} \\ 
  \midrule
        \multirow{4}{*}{Baselines} 
        & Random & .670 & .198 & .032 & .012 & .281 \\
    	& Clippy \cite{Ranasinghe2022PerceptualGI} & .757 & .459  & .263  & .049 & .539 \\
		& GroupViT \cite{Xu2022GroupViTSS} & .862 & .778 & .602 & .205 & .578 \\
		& LSeg \cite{li2022language} & .952 & .897 & .758 & .311 & .678 \\
  \midrule
        \multirow{6}{*}{\begin{tabular}{c}\textbf{Peekaboo}\\Variants (ours)\end{tabular}} 
		& Raster & .756 & .323 & .103 & .012 & .340 \\
		& CLIP & .918 & .488 & .093 & .023 & .430 \\
		& Fourier & .862 & .598 & .231 & .084 & .454 \\
		& Bilateral Fourier & .845 & .608 & .281 & .123 & .470 \\
		& Depth Bilateral & .929 & .707 & .455 & .187 & .551 \\
		& \textbf{RGB Bilateral} & .892 & .709 & .331 & .130 & .520 \\
        \bottomrule
	\end{tabular}
	\caption{\textbf{VOC-C Semantic Segmentation Evaluation.}
	Our results on the VOC-C dataset report mIoU values for our method and GroupViT baseline. Numbers for our baselines are obtained running their respective pre-trained models. Numbers reported are precision and mean-IoU values.}
	\label{tbl:pascal}
	\vspace{-0.5em}
\end{table}

We present evaluations on the VOC-C dataset in \cref{tbl:pascal}. Our modified Pascal VOC dataset, VOC-C, was created by selecting and cropping images  from the original VOC2012 dataset with single subjects larger than 128x128 pixels.

The text prompts VOC-C dataset relatively simple. They're simply the names of the dataset's 20 classes. Some examples: ``cat'', ``dog'', ``aeroplane'', ``bird'' - etc. 

While Peekaboo doesn't outperform Clippy, GroupViT or LSeg, it has fairly competitive performance. In fact, Peekaboo's \vart{Depth Bilateral} variant even outperforms Clippy. And to reiterate, unlike these baselines, Peekaboo requires no training whatsoever.

\subsection{Qualitative Results}


\begin{figure}[h]
\centering
\begin{minipage}{\linewidth}
    \centering
    \includegraphics[width=0.3125\linewidth]{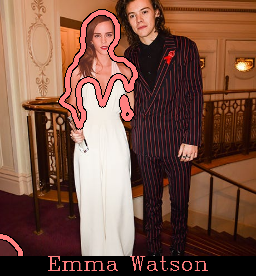}
    \includegraphics[width=0.3125\linewidth]{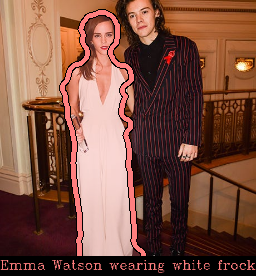}
    \includegraphics[width=0.23125\linewidth]{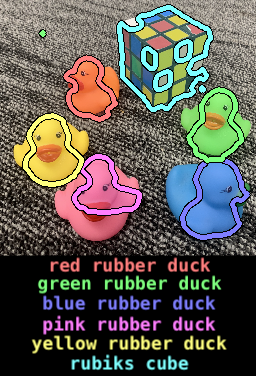}
\end{minipage}

\caption{\textbf{Varying granularity (left)}:
Another feature of \modelname is its ability to segment at differing granularity. While the model performance in such situations is quite sensitive to the caption, we highlight how subtle variations to the caption allows us to localize relevant regions of image accordingly. \textbf{Real-world applications (right)}:
    \modelname can localize real-world objects from arbitrary captions. We highlight how each localization successfully captures the relevant region with a clear overlap over each object centroid. This is particularly useful in applications such as robotics where interaction with arbitrary objects defined by natural language can be valuable.}
\label{fig:vis_ewQ}
\end{figure}

\begin{figure*}[h]

\begin{minipage}{\linewidth}
	\includegraphics[width=0.195\linewidth]{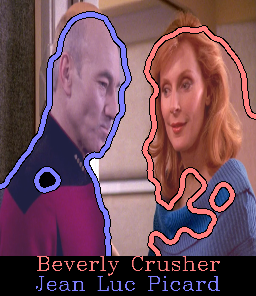}
	\includegraphics[width=0.195\linewidth]{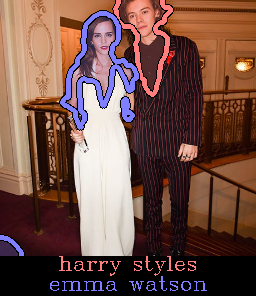}
	\includegraphics[width=0.195\linewidth]{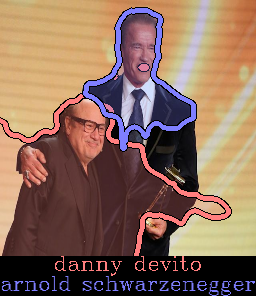}
	\includegraphics[width=0.195\linewidth]{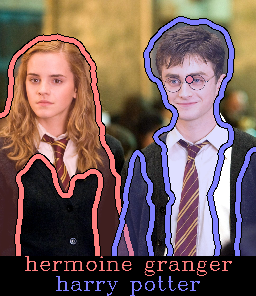}
	\includegraphics[width=0.195\linewidth]{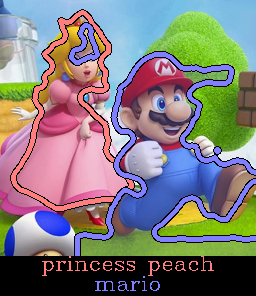}
\end{minipage}
\begin{minipage}{\linewidth}
	\includegraphics[width=0.195\linewidth]{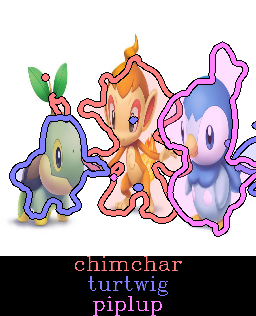}
	\includegraphics[width=0.195\linewidth]{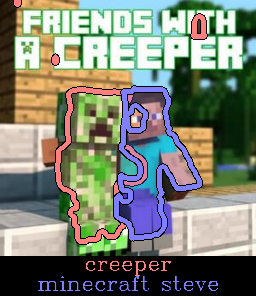}
	\includegraphics[width=0.195\linewidth]{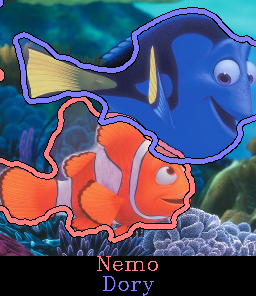}
	\includegraphics[width=0.195\linewidth]{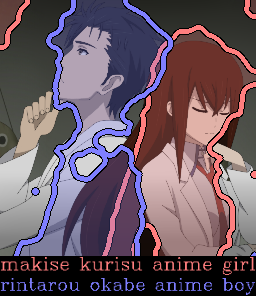}
	\includegraphics[width=0.195\linewidth]{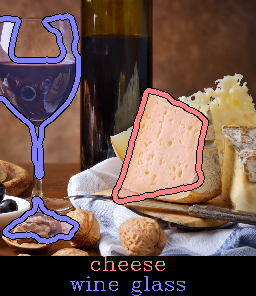}
\end{minipage}
\caption{\textbf{\modelname is truly open vocabulary}:
We illustrate examples where \modelname is able to localize various regions of interest defined by references from popular culture. All of these are examples where the model has been able to correctly localize (identified region centroid or high IoU with correct region). The caption below each image is used to generate the same color mask. 
An interesting behavior of our model is its localization to the region most defined by the accompanying text caption. For example, in the case of Emma Watson in the top row second figure, it is visible how \modelname localizes on her face and body instead of her frock (see \cref{fig:vis_ewQ} for more on this).
}

\label{fig:vis_moreQ}

\end{figure*}

In this section, we present some more qualitative evaluations of our proposed method. In \cref{fig:vis_moreQ}, we show some randomly selected examples where \modelname successfully localizes regions of interest based on popular cultural references. 
We hypothesize that our \modelname is able to understand such a wide vocabulary due to the strength of the pre-trained diffusion model, which was trained on an interet-scale dataset. 
We also analyze real-world image captured by our camera in \cref{fig:vis_moreQ}.


Another characteristic of this open vocabulary nature that endows our model is to probe image regions at varying granularities. We illustrate this behavior in \cref{fig:vis_ewQ} where sub-regions of a single person can be localized based on different text captions.

\section{Generating Images with Transparency}
\label{sec:rgbagen}

Peekaboo loss can do more than just segment images - it can generate images with transparency. In fact, it does this quite reliably - giving detailed alpha masks along with each generated image.
Images with transparency are incredibly useful for many domains, such as graphic design assets and video game textures. However, to the best of our knowledge, all current text-to-image models such  as \cite{imagen}\cite{StableDiffusion}\cite{parti}\cite{dalle2} have trained strictly on RGB images, and are not designed to generate images with an alpha channel.

Despite the absence of such models, we can generate RGBA images by iteratively optimizing an RGB image jointly with an alpha mask using Peekaboo loss. Pseudocode for this process is given in Algorithm \ref{alg:rgbageneration}, which builds on Algorithm \ref{alg:peekaboo}. 

\begin{algorithm}[h]
    \caption{Pytorch style pseudocode for transparent RGBA image generation}
    \label{alg:rgbageneration}
    \definecolor{codeblue}{rgb}{0.25,0.5,0.5}
    \definecolor{codeorange}{rgb}{1.0,0.5,0.3} 
    \definecolor{codegreen}{rgb}{0.13,0.54,0.13}
    \lstset{
      basicstyle=\fontsize{7.2pt}{7.2pt}\ttfamily\bfseries,
      commentstyle=\fontsize{7.2pt}{7.2pt}\color{codeblue},
      keywordstyle=\fontsize{7.2pt}{7.2pt}\color{codeorange},
      stringstyle=\color{codegreen},
      numbers=left,  
      numbersep=5pt, 
      numberstyle=\fontsize{6.2pt}{6.2pt}\color{codeblue},
      columns=fixed,
      xleftmargin=2em,
      framexleftmargin=1.5em
    }
    \begin{lstlisting}[language=python] 
    #Continued from Algorithm 1
    def text_to_rgba_image(prompt):
    	"""Given a text prompt, generate a new RGBA image
    	(An image with an alpha transparency mask)"""
    	
    	alpha = LearnableAlphaMask() # nn.Module 
    	image = LearnableRGBImage() # nn.Module 
    	optim = torch.optim.SGD(alpha.parameters(), image.parameters())
    	
    	for _ in range(num_iterations):
    		peekaboo_loss(image(), prompt, alpha()).backward()
    		optim.step() ; optim.zero_grad()
    	return image(), alpha()
    \end{lstlisting}
\end{algorithm}

\begin{figure}[h]
\centering
	\includegraphics[width=0.99\linewidth]{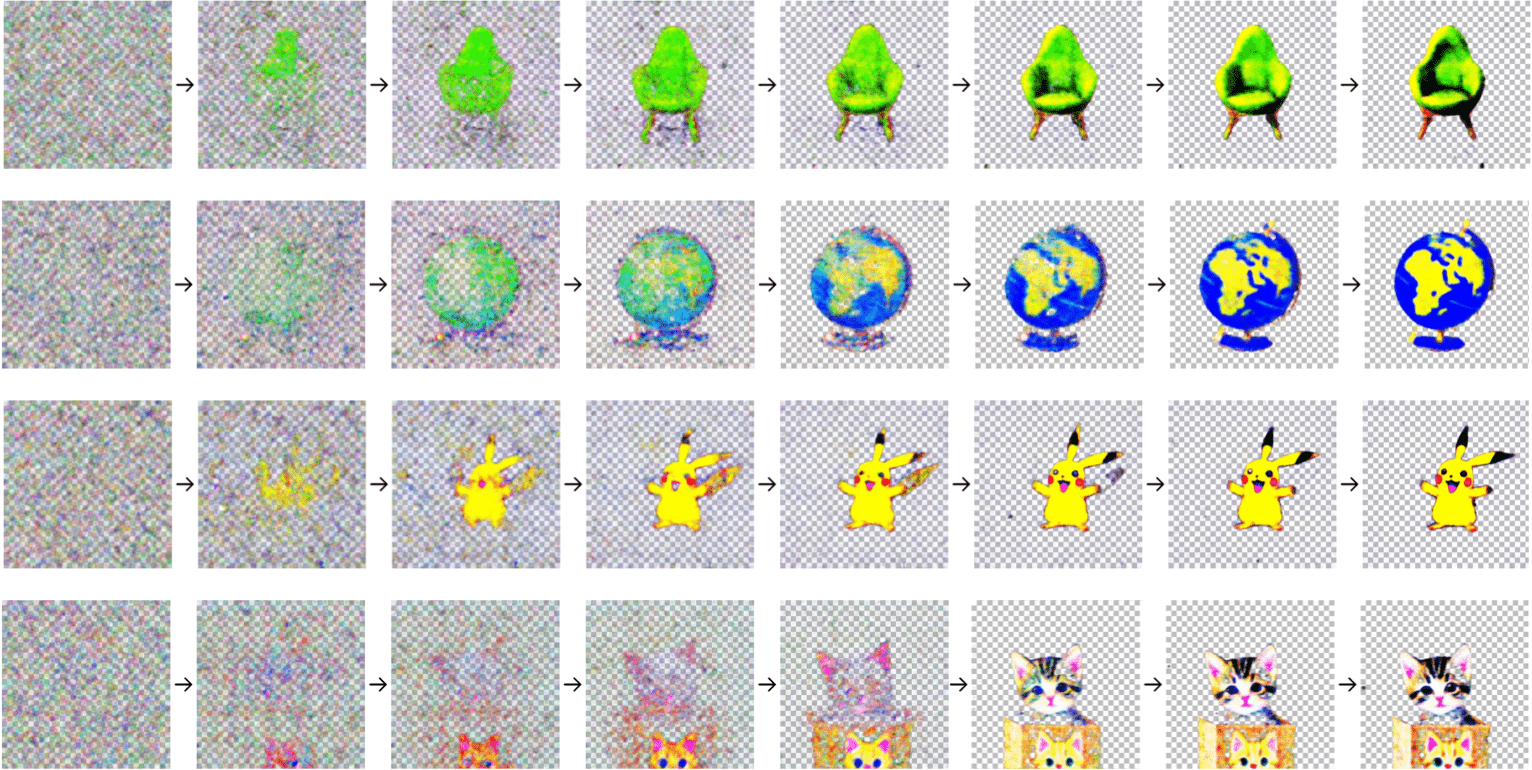}
	\caption{\textbf{Image Generation Timelapse}:
	Above we show a timelapse of the RGBA image generation process. The prompts from top to bottom are ``avocado armchair'', ``globe'', ``pikachu'', ``cat in a box''.  More examples are in the appendix. The images are on  checkerboards to illustrate transparency. 
	}
	\label{fig:rgbatimelapse}
\end{figure}

\begin{figure}[h]
\centering
	\includegraphics[width=0.8\linewidth]{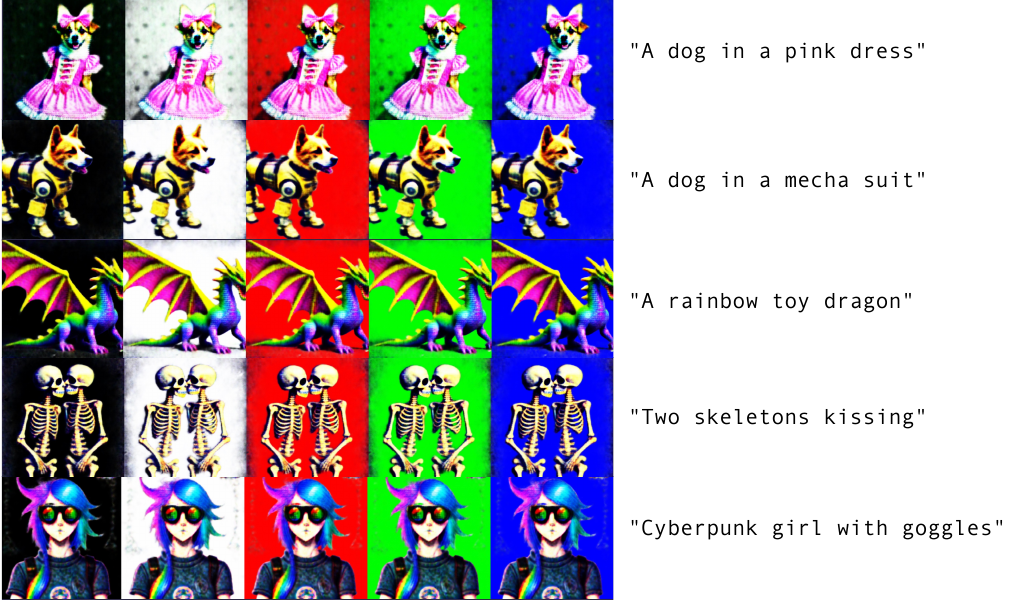}
	\caption{\textbf{Image Generation Examples}:
	Above we show five more examples of RGBA images generated with this method, placed on different backgrounds. Note how Peekaboo successfully separates high-frequency details, for example, between rib bones and the background. 
	}
	\label{fig:rgbaexamples}
\end{figure}

Example images can be found in \cref{fig:rgbaexamples} and \cref{fig:rgbatimelapse}.

This process of creating images via optimization is very similar to \cite{burgert2023diffusion_illusions}, which also uses score distillation loss to optimize 2d images. However, it also suffers from some of the same caveats.

\section{Conclusion}
\label{sec:conclusion}
In this work, we established how text-to-image diffusion models trained on large-scale internet data contain strong cues on localization. Note that their training process contains no explicit modelling of localization. We then propose a novel \textit{inference-time optimization} technique that can extract this localization knowledge and apply this to downstream referring and semantic segmentation tasks. In particular, we perform such segmentation with no segmentation-specific re-training of the diffusion modell, leaving its weights (and therein all strengths of the model) intact.

The major limitation of our work is its reliance on a large diffusion model pre-trained on internet-scale data. Given the difficulty of training such diffusion models under usual academic settings (resource constraints), approaches building off \modelname must rely on publicly available diffusion models. Moreover, all flaws and biases contained within such diffusion models \cite{sueing_sd} as well as internet-scale datasets used for training \cite{birhane2021multimodal} will be transferred to \modelname.


\newpage
{\small
\bibliographystyle{unsrt}
\bibliography{egbib}

\begin{thebibliography}{10}

\bibitem{coco}
Tsung-Yi Lin, Michael Maire, Serge Belongie, James Hays, Pietro Perona, Deva
  Ramanan, Piotr Doll{\'a}r, and C~Lawrence Zitnick.
\newblock Microsoft coco: Common objects in context.
\newblock In {\em ECCV}, 2014.

\bibitem{hu2016segmentation}
Ronghang Hu, Marcus Rohrbach, and Trevor Darrell.
\newblock Segmentation from natural language expressions.
\newblock In {\em ECCV}, 2016.

\bibitem{sun2020scalability}
Pei Sun, Henrik Kretzschmar, Xerxes Dotiwalla, Aurelien Chouard, Vijaysai
  Patnaik, Paul Tsui, James Guo, Yin Zhou, Yuning Chai, Benjamin Caine, et~al.
\newblock Scalability in perception for autonomous driving: Waymo open dataset.
\newblock In {\em CVPR}, pages 2446--2454, 2020.

\bibitem{chen2017deeplab}
Liang-Chieh Chen, George Papandreou, Iasonas Kokkinos, Kevin Murphy, and Alan~L
  Yuille.
\newblock Deeplab: Semantic image segmentation with deep convolutional nets,
  atrous convolution, and fully connected crfs.
\newblock {\em IEEE TPAMI}, 2017.

\bibitem{chen2018searching}
Liang-Chieh Chen, Maxwell Collins, Yukun Zhu, George Papandreou, Barret Zoph,
  Florian Schroff, Hartwig Adam, and Jonathon Shlens.
\newblock Searching for efficient multi-scale architectures for dense image
  prediction.
\newblock {\em NeurIPS}, 31, 2018.

\bibitem{Ghiasi2021OpenVocabularyIS}
Golnaz Ghiasi, Xiuye Gu, Yin Cui, and Tsung-Yi Lin.
\newblock Open-vocabulary image segmentation.
\newblock {\em ArXiv}, abs/2112.12143, 2021.

\bibitem{li2022language}
Boyi Li, Kilian~Q Weinberger, Serge Belongie, Vladlen Koltun, and Ren{\'e}
  Ranftl.
\newblock Language-driven semantic segmentation.
\newblock {\em ICLR}, 2022.

\bibitem{clip}
Alec Radford, Jong~Wook Kim, Chris Hallacy, Aditya Ramesh, Gabriel Goh,
  Sandhini Agarwal, Girish Sastry, Amanda Askell, Pamela Mishkin, Jack Clark,
  et~al.
\newblock Learning transferable visual models from natural language
  supervision.
\newblock {\em ICML}, 2021.

\bibitem{jia2021scaling}
Chao Jia, Yinfei Yang, Ye~Xia, Yi-Ting Chen, Zarana Parekh, Hieu Pham, Quoc~V
  Le, Yunhsuan Sung, Zhen Li, and Tom Duerig.
\newblock Scaling up visual and vision-language representation learning with
  noisy text supervision.
\newblock {\em Int. Conf. on Mach. Learn.}, 2021.

\bibitem{Wang2022CRISCR}
Zhaoqing Wang, Yu~Lu, Qiang Li, Xunqiang Tao, Yan Guo, Ming Gong, and Tongliang
  Liu.
\newblock Cris: Clip-driven referring image segmentation.
\newblock {\em 2022 IEEE/CVF Conference on Computer Vision and Pattern
  Recognition (CVPR)}, pages 11676--11685, 2022.

\bibitem{Xu2022GroupViTSS}
Jiarui Xu, Shalini~De Mello, Sifei Liu, Wonmin Byeon, Thomas Breuel, Jan Kautz,
  and X.~Wang.
\newblock Groupvit: Semantic segmentation emerges from text supervision.
\newblock {\em 2022 IEEE/CVF Conference on Computer Vision and Pattern
  Recognition (CVPR)}, pages 18113--18123, 2022.

\bibitem{Ranasinghe2022PerceptualGI}
Kanchana Ranasinghe, Brandon McKinzie, Sachin Ravi, Yinfei Yang, Alexander
  Toshev, and Jonathon Shlens.
\newblock Perceptual grouping in vision-language models.
\newblock {\em ArXiv}, abs/2210.09996, 2022.

\bibitem{ddpm}
Jonathan Ho, Ajay Jain, and Pieter Abbeel.
\newblock Denoising diffusion probabilistic models.
\newblock {\em NeurIPS}, 2020.

\bibitem{pmlr-v37-sohl-dickstein15}
Jascha Sohl-Dickstein, Eric Weiss, Niru Maheswaranathan, and Surya Ganguli.
\newblock Deep unsupervised learning using nonequilibrium thermodynamics.
\newblock {\em ICML}, 2015.

\bibitem{scoresde}
Yang Song, Jascha Sohl-Dickstein, Diederik~P Kingma, Abhishek Kumar, Stefano
  Ermon, and Ben Poole.
\newblock Score-based generative modeling through stochastic differential
  equations.
\newblock {\em ICLR}, 2021.

\bibitem{ODISE}
Jiarui Xu, Sifei Liu, Arash Vahdat, Wonmin Byeon, Xiaolong Wang, and Shalini~De
  Mello.
\newblock Open-vocabulary panoptic segmentation with text-to-image diffusion
  models, 2023.

\bibitem{Rombach2022HighResolutionIS}
Robin Rombach, A.~Blattmann, Dominik Lorenz, Patrick Esser, and Bj{\"o}rn
  Ommer.
\newblock High-resolution image synthesis with latent diffusion models.
\newblock {\em 2022 IEEE/CVF Conference on Computer Vision and Pattern
  Recognition (CVPR)}, pages 10674--10685, 2022.

\bibitem{Kazemzadeh2014ReferItGameRT}
Sahar Kazemzadeh, Vicente Ordonez, Marc andre Matten, and Tamara~L. Berg.
\newblock Referitgame: Referring to objects in photographs of natural scenes.
\newblock In {\em EMNLP}, 2014.

\bibitem{everingham2010pascal}
Mark Everingham, Luc Van~Gool, Christopher~KI Williams, John Winn, and Andrew
  Zisserman.
\newblock The pascal visual object classes (voc) challenge.
\newblock {\em IJCV}, 2010.

\bibitem{frome2013devise}
Andrea Frome, Greg~S Corrado, Jonathon Shlens, Samy Bengio, Jeff Dean,
  Marc'Aurelio Ranzato, and Tomas Mikolov.
\newblock Devise: A deep visual-semantic embedding model.
\newblock {\em NeurIPS}, 26, 2013.

\bibitem{socher2013zero}
Richard Socher, Milind Ganjoo, Christopher~D Manning, and Andrew Ng.
\newblock Zero-shot learning through cross-modal transfer.
\newblock {\em NeurIPS}, 26, 2013.

\bibitem{karpathy2015deep}
Andrej Karpathy and Li~Fei-Fei.
\newblock Deep visual-semantic alignments for generating image descriptions.
\newblock In {\em CVPR}, pages 3128--3137, 2015.

\bibitem{vinyals2015show}
Oriol Vinyals, Alexander Toshev, Samy Bengio, and Dumitru Erhan.
\newblock Show and tell: A neural image caption generator.
\newblock In {\em CVPR}, pages 3156--3164, 2015.

\bibitem{kiros2014unifying}
Ryan Kiros, Ruslan Salakhutdinov, and Richard~S Zemel.
\newblock Unifying visual-semantic embeddings with multimodal neural language
  models.
\newblock {\em arXiv preprint arXiv:1411.2539}, 2014.

\bibitem{mao2014explain}
Junhua Mao, Wei Xu, Yi~Yang, Jiang Wang, and Alan~L Yuille.
\newblock Explain images with multimodal recurrent neural networks.
\newblock {\em arXiv preprint arXiv:1410.1090}, 2014.

\bibitem{pham2021combined}
Hieu Pham, Zihang Dai, Golnaz Ghiasi, Hanxiao Liu, Adams~Wei Yu, Minh-Thang
  Luong, Mingxing Tan, and Quoc~V Le.
\newblock Combined scaling for zero-shot transfer learning.
\newblock {\em arXiv preprint arXiv:2111.10050}, 2021.

\bibitem{yu2022coca}
Jiahui Yu, Zirui Wang, Vijay Vasudevan, Legg Yeung, Mojtaba Seyedhosseini, and
  Yonghui Wu.
\newblock Coca: Contrastive captioners are image-text foundation models.
\newblock {\em arXiv preprint arXiv:2205.01917}, 2022.

\bibitem{desai2021virtex}
Karan Desai and Justin Johnson.
\newblock Virtex: Learning visual representations from textual annotations.
\newblock In {\em CVPR}, pages 11162--11173, 2021.

\bibitem{yao2022filip}
Lewei Yao, Runhui Huang, Lu~Hou, Guansong Lu, Minzhe Niu, Hang Xu, Xiaodan
  Liang, Zhenguo Li, Xin Jiang, and Chunjing Xu.
\newblock {FILIP}: Fine-grained interactive language-image pre-training.
\newblock In {\em ICLR}, 2022.

\bibitem{cui2022democratizing}
Yufeng Cui, Lichen Zhao, Feng Liang, Yangguang Li, and Jing Shao.
\newblock Democratizing contrastive language-image pre-training: A clip
  benchmark of data, model, and supervision.
\newblock {\em arXiv preprint arXiv:2203.05796}, 2022.

\bibitem{Zeng2022SocraticMC}
Andy Zeng, Adrian~S. Wong, Stefan Welker, Krzysztof Choromanski, Federico
  Tombari, Aveek Purohit, Michael~S. Ryoo, Vikas Sindhwani, Johnny Lee, Vincent
  Vanhoucke, and Peter~R. Florence.
\newblock Socratic models: Composing zero-shot multimodal reasoning with
  language.
\newblock {\em ArXiv}, abs/2204.00598, 2022.

\bibitem{yuan2021florence}
Lu~Yuan, Dongdong Chen, Yi-Ling Chen, Noel Codella, Xiyang Dai, Jianfeng Gao,
  Houdong Hu, Xuedong Huang, Boxin Li, Chunyuan Li, et~al.
\newblock Florence: A new foundation model for computer vision.
\newblock {\em arXiv preprint arXiv:2111.11432}, 2021.

\bibitem{kamath2021mdetr}
Aishwarya Kamath, Mannat Singh, Yann LeCun, Gabriel Synnaeve, Ishan Misra, and
  Nicolas Carion.
\newblock Mdetr-modulated detection for end-to-end multi-modal understanding.
\newblock In {\em ICCV}, pages 1780--1790, 2021.

\bibitem{Gu2022OpenvocabularyOD}
Xiuye Gu, Tsung-Yi Lin, Weicheng Kuo, and Yin Cui.
\newblock Open-vocabulary object detection via vision and language knowledge
  distillation.
\newblock In {\em ICLR}, 2022.

\bibitem{Nichol2022GLIDETP}
Alex Nichol, Prafulla Dhariwal, Aditya Ramesh, Pranav Shyam, Pamela Mishkin,
  Bob McGrew, Ilya Sutskever, and Mark Chen.
\newblock Glide: Towards photorealistic image generation and editing with
  text-guided diffusion models.
\newblock In {\em ICML}, 2022.

\bibitem{dalle}
Aditya Ramesh, Mikhail Pavlov, Gabriel Goh, Scott Gray, Chelsea Voss, Alec
  Radford, Mark Chen, and Ilya Sutskever.
\newblock Zero-shot text-to-image generation.
\newblock {\em ICML}, 2021.

\bibitem{dalle2}
Aditya Ramesh, Prafulla Dhariwal, Alex Nichol, Casey Chu, and Mark Chen.
\newblock Hierarchical text-conditional image generation with clip latents,
  2022.

\bibitem{imagen}
Chitwan Saharia, William Chan, Saurabh Saxena, Lala Li, Jay Whang, Emily
  Denton, Seyed Kamyar~Seyed Ghasemipour, Burcu~Karagol Ayan, S.~Sara Mahdavi,
  Rapha~Gontijo Lopes, Tim Salimans, Jonathan Ho, David~J Fleet, and Mohammad
  Norouzi.
\newblock Photorealistic text-to-image diffusion models with deep language
  understanding.
\newblock {\em arXiv:2205.11487}, 2022.

\bibitem{palette}
Chitwan Saharia, William Chan, Huiwen Chang, Chris~A. Lee, Jonathan Ho, Tim
  Salimans, David~J. Fleet, and Mohammad Norouzi.
\newblock Palette: Image-to-image diffusion models, 2021.

\bibitem{parti}
Jiahui Yu, Yuanzhong Xu, Jing~Yu Koh, Thang Luong, Gunjan Baid, Zirui Wang,
  Vijay Vasudevan, Alexander Ku, Yinfei Yang, Burcu~Karagol Ayan, Ben
  Hutchinson, Wei Han, Zarana Parekh, Xin Li, Han Zhang, Jason Baldridge, and
  Yonghui Wu.
\newblock Scaling autoregressive models for content-rich text-to-image
  generation.
\newblock {\em arXiv:2206.10789}, 2022.

\bibitem{sr3}
Chitwan Saharia, Jonathan Ho, William Chan, Tim Salimans, David~J. Fleet, and
  Mohammad Norouzi.
\newblock Image super-resolution via iterative refinement, 2021.

\bibitem{wavegrad}
Nanxin Chen, Yu~Zhang, Heiga Zen, Ron~J Weiss, Mohammad Norouzi, and William
  Chan.
\newblock Wavegrad: Estimating gradients for waveform generation.
\newblock {\em arXiv:2009.00713}, 2020.

\bibitem{videodiffusion}
Jonathan Ho, Tim Salimans, Alexey Gritsenko, William Chan, Mohammad Norouzi,
  and David~J Fleet.
\newblock Video diffusion models.
\newblock {\em arXiv:2204.03458}, 2022.

\bibitem{diffwave}
Zhifeng Kong, Wei Ping, Jiaji Huang, Kexin Zhao, and Bryan Catanzaro.
\newblock Diffwave: A versatile diffusion model for audio synthesis.
\newblock {\em ICLR}, 2021.

\bibitem{Poole2022DreamFusionTU}
Ben Poole, Ajay Jain, Jonathan~T. Barron, and Ben Mildenhall.
\newblock Dreamfusion: Text-to-3d using 2d diffusion.
\newblock {\em ArXiv}, abs/2209.14988, 2022.

\bibitem{graikos2022diffusion}
Alexandros Graikos, Nikolay Malkin, Nebojsa Jojic, and Dimitris Samaras.
\newblock Diffusion models as plug-and-play priors.
\newblock In {\em Thirty-Sixth Conference on Neural Information Processing
  Systems}, 2022.

\bibitem{mildenhall2020nerf}
Ben Mildenhall, Pratul~P. Srinivasan, Matthew Tancik, Jonathan~T. Barron, Ravi
  Ramamoorthi, and Ren Ng.
\newblock {NeRF}: Representing scenes as neural radiance fields for view
  synthesis.
\newblock {\em ECCV}, 2020.

\bibitem{crowson2022vqganclip}
Katherine Crowson, Stella Biderman, Daniel Kornis, Dashiell Stander, Eric
  Hallahan, Louis Castricato, and Edward Raff.
\newblock Vqgan-clip: Open domain image generation and editing with natural
  language guidance, 2022.

\bibitem{jain2021dreamfields}
Ajay Jain, Ben Mildenhall, Jonathan~T. Barron, Pieter Abbeel, and Ben Poole.
\newblock Zero-shot text-guided object generation with dream fields.
\newblock {\em CVPR}, 2022.

\bibitem{khalid2022clipmesh}
Nasir~Mohammad Khalid, Tianhao Xie, Eugene Belilovsky, and Tiberiu Popa.
\newblock Clip-mesh: Generating textured meshes from text using pretrained
  image-text models, 2022.

\bibitem{burgert2023diffusion_illusions}
Ryan Burgert, Kanchana Ranasinghe, Xiang Li, and Michael Ryoo.
\newblock Diffusion illusions: Hiding images in plain sight.
\newblock \url{https://ryanndagreat.github.io/Diffusion-Illusions}, March 2023.
\newblock Accessed: 2023-03-17.

\bibitem{lin2022magic3d}
Chen-Hsuan Lin, Jun Gao, Luming Tang, Towaki Takikawa, Xiaohui Zeng, Xun Huang,
  Karsten Kreis, Sanja Fidler, Ming-Yu Liu, and Tsung-Yi Lin.
\newblock Magic3d: High-resolution text-to-3d content creation.
\newblock {\em arXiv preprint arXiv:2211.10440}, 2022.

\bibitem{wordasimage}
Shir Iluz, Yael Vinker, Amir Hertz, Daniel Berio, Daniel Cohen-Or, and Ariel
  Shamir.
\newblock Word-as-image for semantic typography.
\newblock \url{https://arxiv.org/abs/2303.01818}, 2023.

\bibitem{malik2001visual}
Jitendra Malik.
\newblock Visual grouping and object recognition.
\newblock In {\em Proceedings 11th International Conference on Image Analysis
  and Processing}, pages 612--621. IEEE, 2001.

\bibitem{comaniciu1997robust}
Dorin Comaniciu and Peter Meer.
\newblock Robust analysis of feature spaces: Color image segmentation.
\newblock In {\em CVPR}, pages 750--755. IEEE, 1997.

\bibitem{shi2000normalized}
Jianbo Shi and Jitendra Malik.
\newblock Normalized cuts and image segmentation.
\newblock {\em IEEE TPAMI}, 22(8):888--905, 2000.

\bibitem{ren2003learning}
Xiaofeng Ren and Jitendra Malik.
\newblock Learning a classification model for segmentation.
\newblock In {\em CVPR}, volume~2, pages 10--10. IEEE Computer Society, 2003.

\bibitem{caron2021emerging}
Mathilde Caron, Hugo Touvron, Ishan Misra, Herv{\'e} J{\'e}gou, Julien Mairal,
  Piotr Bojanowski, and Armand Joulin.
\newblock Emerging properties in self-supervised vision transformers.
\newblock In {\em CVPR}, pages 9650--9660, 2021.

\bibitem{hamilton2022unsupervised}
Mark Hamilton, Zhoutong Zhang, Bharath Hariharan, Noah Snavely, and William~T
  Freeman.
\newblock Unsupervised semantic segmentation by distilling feature
  correspondences.
\newblock {\em ICLR}, 2022.

\bibitem{Cho2021PiCIEUS}
Jang~Hyun Cho, Utkarsh Mall, Kavita Bala, and Bharath Hariharan.
\newblock Picie: Unsupervised semantic segmentation using invariance and
  equivariance in clustering.
\newblock {\em CVPR}, pages 16789--16799, 2021.

\bibitem{VanGansbeke2021UnsupervisedSS}
Wouter~Van Gansbeke, Simon Vandenhende, Stamatios Georgoulis, and Luc~Van Gool.
\newblock Unsupervised semantic segmentation by contrasting object mask
  proposals.
\newblock {\em ICCV}, pages 10032--10042, 2021.

\bibitem{Ji2019InvariantIC}
Xu~Ji, Andrea Vedaldi, and Jo{\~a}o~F. Henriques.
\newblock Invariant information clustering for unsupervised image
  classification and segmentation.
\newblock {\em ICCV}, pages 9864--9873, 2019.

\bibitem{yu2018mattnet}
Licheng Yu, Zhe Lin, Xiaohui Shen, Jimei Yang, Xin Lu, Mohit Bansal, and
  Tamara~L Berg.
\newblock {MAttNet}: Modular attention network for referring expression
  comprehension.
\newblock In {\em CVPR}, 2018.

\bibitem{ye2019cross}
Linwei Ye, Mrigank Rochan, Zhi Liu, and Yang Wang.
\newblock Cross-modal self-attention network for referring image segmentation.
\newblock In {\em CVPR}, 2019.

\bibitem{liu2017recurrent}
Chenxi Liu, Zhe Lin, Xiaohui Shen, Jimei Yang, Xin Lu, and Alan Yuille.
\newblock Recurrent multimodal interaction for referring image segmentation.
\newblock In {\em ICCV}, 2017.

\bibitem{li2018referring}
Ruiyu Li, Kaican Li, Yi-Chun Kuo, Michelle Shu, Xiaojuan Qi, Xiaoyong Shen, and
  Jiaya Jia.
\newblock Referring image segmentation via recurrent refinement networks.
\newblock In {\em CVPR}, 2018.

\bibitem{margffoy2018dynamic}
Edgar Margffoy-Tuay, Juan~C P{\'e}rez, Emilio Botero, and Pablo Arbel{\'a}ez.
\newblock Dynamic multimodal instance segmentation guided by natural language
  queries.
\newblock In {\em ECCV}, 2018.

\bibitem{chen2019referring}
Yi~Wen Chen, Yi~Hsuan Tsai, Tiantian Wang, Yen~Yu Lin, and Ming~Hsuan Yang.
\newblock Referring expression object segmentation with caption-aware
  consistency.
\newblock In {\em BMVC}, 2019.

\bibitem{shi2018key}
Hengcan Shi, Hongliang Li, Fanman Meng, and Qingbo Wu.
\newblock Key-word-aware network for referring expression image segmentation.
\newblock In {\em ECCV}, 2018.

\bibitem{huang2020referring}
Shaofei Huang, Tianrui Hui, Si~Liu, Guanbin Li, Yunchao Wei, Jizhong Han, Luoqi
  Liu, and Bo~Li.
\newblock Referring image segmentation via cross-modal progressive
  comprehension.
\newblock In {\em CVPR}, 2020.

\bibitem{huilinguistic}
Tianrui Hui, Si~Liu, Shaofei Huang, Guanbin Li, Sansi Yu, Faxi Zhang, and
  Jizhong Han.
\newblock Linguistic structure guided context modeling for referring image
  segmentation.
\newblock In {\em ECCV}, 2020.

\bibitem{SAM}
Alexander Kirillov, Eric Mintun, Nikhila Ravi, Hanzi Mao, Chloe Rolland, Laura
  Gustafson, Tete Xiao, Spencer Whitehead, Alexander~C. Berg, Wan-Yen Lo, Piotr
  Doll{\'a}r, and Ross Girshick.
\newblock Segment anything.
\newblock {\em arXiv:2304.02643}, 2023.

\bibitem{SEEM}
Shilong Liu, Zhaoyang Zeng, Tianhe Ren, Feng Li, Hao Zhang, Jie Yang, Chunyuan
  Li, Jianwei Yang, Hang Su, Jun Zhu, et~al.
\newblock Grounding dino: Marrying dino with grounded pre-training for open-set
  object detection.
\newblock {\em arXiv preprint arXiv:2303.05499}, 2023.

\bibitem{ffn}
Matthew Tancik, Pratul~P. Srinivasan, Ben Mildenhall, Sara Fridovich-Keil,
  Nithin Raghavan, Utkarsh Singhal, Ravi Ramamoorthi, Jonathan~T. Barron, and
  Ren Ng.
\newblock Fourier features let networks learn high frequency functions in low
  dimensional domains.
\newblock {\em NeurIPS}, 2020.

\bibitem{Burgert2022}
Ryan Burgert, Jinghuan Shang, Xiang Li, and Michael Ryoo.
\newblock Neural neural textures make sim2real consistent.
\newblock In {\em Proceedings of the 6th Conference on Robot Learning}, 2022.

\bibitem{midas}
Ren\'{e} Ranftl, Katrin Lasinger, David Hafner, Konrad Schindler, and Vladlen
  Koltun.
\newblock Towards robust monocular depth estimation: Mixing datasets for
  zero-shot cross-dataset transfer.
\newblock {\em IEEE Transactions on Pattern Analysis and Machine Intelligence},
  44(3), 2022.

\bibitem{StableDiffusion}
Robin Rombach, A.~Blattmann, Dominik Lorenz, Patrick Esser, and Bj{\"o}rn
  Ommer.
\newblock High-resolution image synthesis with latent diffusion models.
\newblock {\em 2022 IEEE/CVF Conference on Computer Vision and Pattern
  Recognition (CVPR)}, pages 10674--10685, 2022.

\bibitem{sueing_sd}
James Vincent.
\newblock Getty images is suing the creators of ai art tool stable diffusion
  for scraping its content.
\newblock
  \url{https://www.theverge.com/2023/1/17/23558516/ai-art-copyright-stable-diffusion-getty-images-lawsuit},
  2023.
\newblock Accessed: 2023-03-17.

\bibitem{birhane2021multimodal}
Abeba Birhane, Vinay~Uday Prabhu, and Emmanuel Kahembwe.
\newblock Multimodal datasets: misogyny, pornography, and malignant
  stereotypes.
\newblock {\em arXiv preprint arXiv:2110.01963}, 2021.

\bibitem{fourier_feature_networks}
Matthew Tancik, Pratul~P. Srinivasan, Ben Mildenhall, Sara Fridovich-Keil,
  Nithin Raghavan, Utkarsh Singhal, Ravi Ramamoorthi, Jonathan~T. Barron, and
  Ren Ng.
\newblock Fourier features let networks learn high frequency functions in low
  dimensional domains.
\newblock {\em NeurIPS}, 2020.

\bibitem{laion5b}
Christoph Schuhmann, Romain Beaumont, Cade~W Gordon, Ross Wightman, mehdi
  cherti, Theo Coombes, Aarush Katta, Clayton Mullis, Patrick Schramowski,
  Srivatsa~R Kundurthy, Katherine Crowson, Richard Vencu, Ludwig Schmidt,
  Robert Kaczmarczyk, and Jenia Jitsev.
\newblock {LAION}-5b: An open large-scale dataset for training next generation
  image-text models.
\newblock {\em NeurIPS Datasets and Benchmarks Track}, 2022.

\end{thebibliography}
}

\newpage
\appendix

\begin{center}

\centering
\vspace{1em}
\textsc{\large \mbox{Peekaboo: Text to Image Diffusion Models are Zero-Shot Segmentors}}
\vspace{3em}
\end{center}

\section{Intuition}
In this section, we aim to convey some intuition behind Peekaboo. We explore the reasoning behind each component used and 
explain
why \modelname works as it does. The key underlying idea is deriving gradients from a conditional diffusion model to guide a mask generation process. 

The diffusion model conditioned on a text caption processes a noisy input to generate a gradient that moves the input toward the target image. This gradient is what we use to guide our learnable alpha mask.
We use Stable Diffusion \cite{Rombach2022HighResolutionIS}, a text-to-image diffusion model trained on billions of internet images. 
This means the model will target to generate images from that image distribution relevant to a text caption. 
The noisy input is the image to be segmented alpha blended with a uniform background. 
Iteratively, it will attempt to update this input to a target distribution image relevant to the text caption. 
Updating the regions of the image most relevant to the text caption makes more sense.
Thus, regions of the image relevant to this text caption will have stronger gradients than regions irrelevant to a prompt. 
For example, consider an image of a dog sitting on a couch. When prompted for ``a dog'', the couch is irrelevant to the prompt and has fewer gradients focused on that region. There is less incentive to remove the alpha mask in that location, and without incentive, it defaults to null due to our alpha regularization loss (\cref{subsec:aux_loss}). This results in the alpha mask focusing on the dog region. 

While this results in our desirable behavior, we observe that it converges to the region most relevant to the text prompt. For example in \cref{fig:vis_ewQ}, in segmenting Emma Watson, it attempts to focus on the region that mostly makes the image look like her, which could be a sub-portion of the human region (in accordance with the accepted notion of segmenting a human).
While Harry Styles could wear a dress, he would still remain Harry Styles. However, only Emma Watson can have her face. Therefore, Emma Watson's face is more essential to the prompt \texttt{Emma Watson}; hence it prioritized segmenting that region while ignoring the dress. 
In a way, one could view this as a new definition of segmentation; \modelname localizes the essence of an object described by language.

\section{Ablations}
In this section, we present some ablations on components of our inference time optimization.

\subsection{Alpha Regularization}
    \newcommand{\gravco}{\lambda_\alpha} 

    The alpha regularization term $\gravity$ (see \cref{sec:alpha_reg}) is important, because it prevents Peekaboo from including irrelevant parts of the image in the alpha mask. By adding this alpha penalty, we effectively tell Peekaboo to create a \textit{minimal} alpha mask. In practice, the $\gravity$ term should be scaled by a constant, which in our experiments was $\gravco = .05$ (the alpha regularization coefficient). 
    
    In this section, we perform ablations where we vary this $\gravco$ term. We play around with the alpha regularization coefficient $\gravco$, scaling it by different factors. For example, ``2x'' means a $\gravco$ that is ``2x'' the normal value, which is $.05 \times 2 = .1$. 

\begin{figure*}[!h]
        \centering
        
        \includegraphics[width=.9\textwidth]{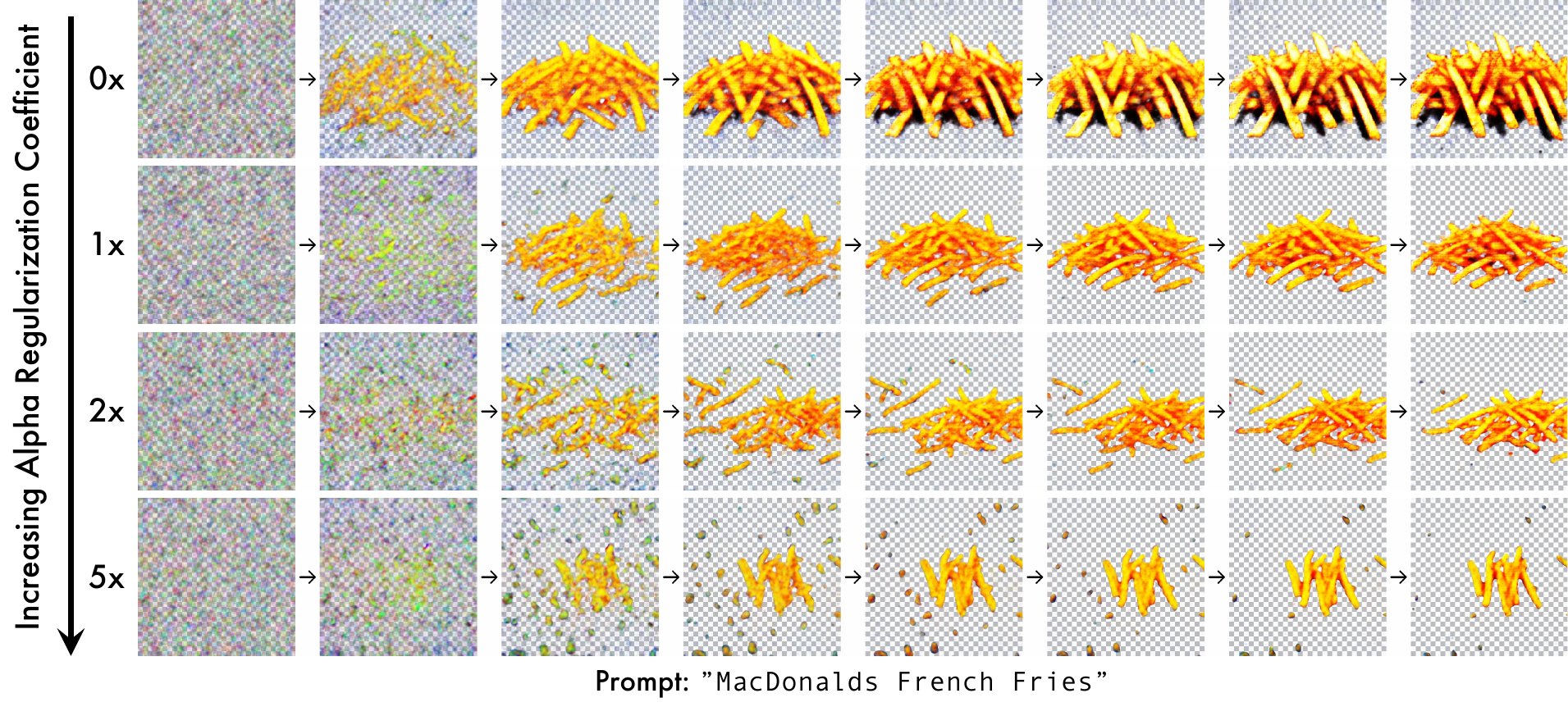}
        \caption{We display four transparent-image generation timelapses for the prompt ``MacDonalds French Fries'' with varying amonuts of alpha regularization. As we increase alpha regularization, Peekaboo tends to generate images with less alpha. In this case, it means less french fries will be generated. Conversely, when the $\gravco$ is eliminated (aka $\gravco = 0$), more french fries than normal are generated.
     }
        \label{fig:alpharegfries}
    \end{figure*}

\begin{figure*}[!h]
        \centering
        
        \includegraphics[width=.9\textwidth]{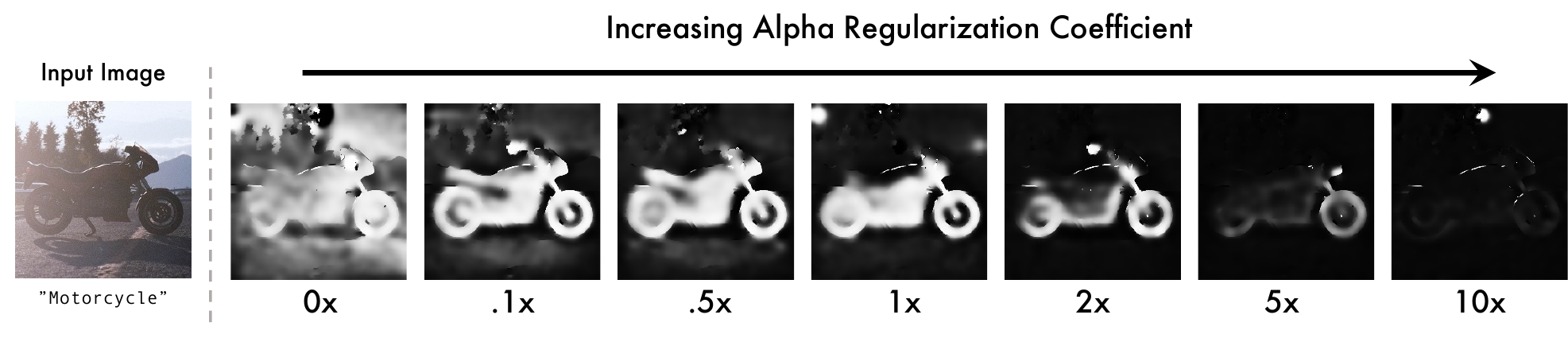}
        \caption{The alpha regularization term plays a large role when using Peekaboo to segment images. If $\gravco$ is too large (on the far right), the entire image might be ignored. Conversely, if $\gravco$ is too small, tons of background details are included.
     }
        \label{fig:alpharegseg}
    \end{figure*}

    We can view the results visually in both the transparent image generation and segmentation contexts. In the transparent image generation task \cref{fig:alpharegfries}, we see that the generated images are more transparent when $\gravco$ is large. Likewise, in the segmentation task \cref{fig:alpharegseg} we can see that Peekaboo generates smaller masks when $\gravco$ is high.
    
\subsection{Implicit neural representations}
\begin{figure*}[!h]
\centering
\begin{minipage}{\linewidth}
    \includegraphics[width=\linewidth]{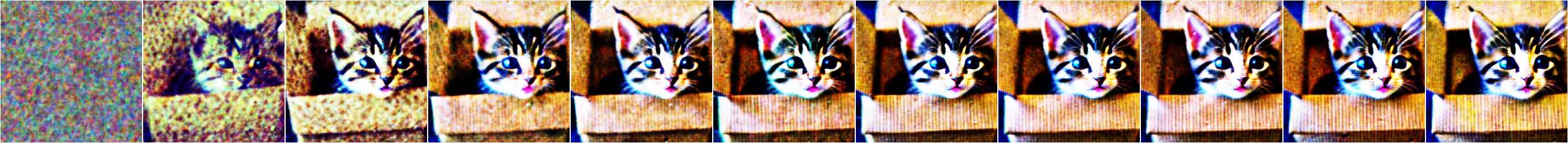}\\
    \includegraphics[width=\linewidth]{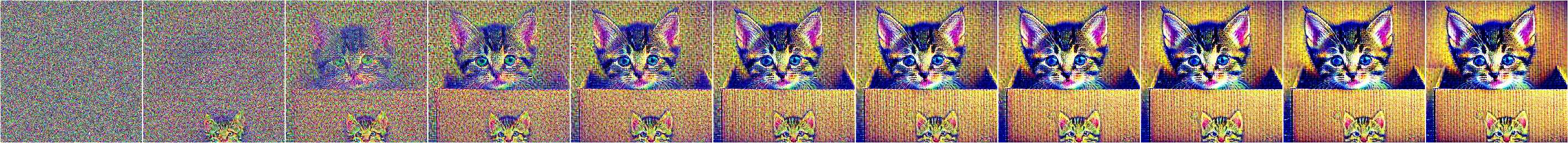} \\
   \hspace*{\fill}\small{$\overrightarrow{\texttt{Iteration}}$}\hspace*{\fill} \\
\end{minipage}
\begin{minipage}{0.5\linewidth}
    \includegraphics[width=0.49\linewidth]{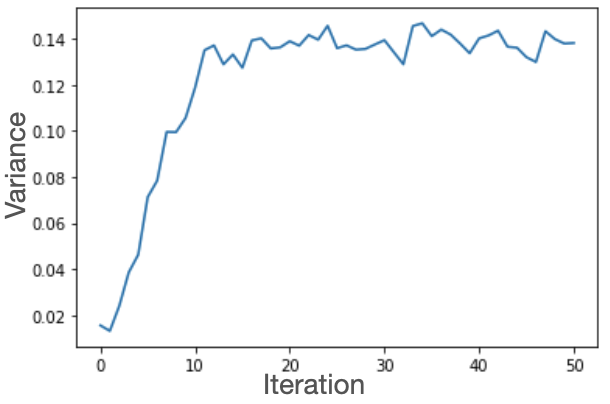}
    \includegraphics[width=0.49\linewidth]{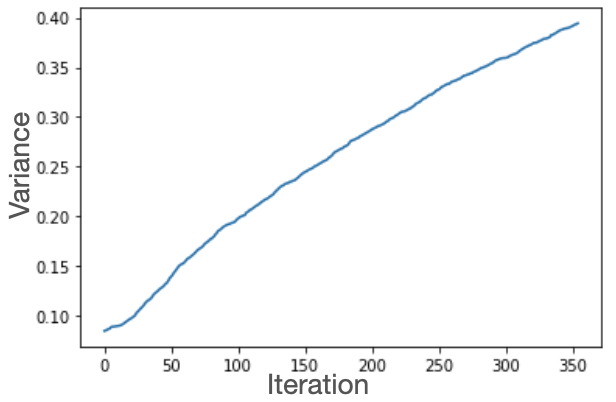}
\end{minipage}
\begin{minipage}{0.49\linewidth}
    \caption{\textbf{Ablation on implicit neural representation}:
    In \modelname's \vart{Fourier} and \vart{Bilateral Fourier} variants, as well as our transparency image generations, we utilize neural-neural textures \cite{Burgert2022}, which use Fourier feature networks \cite{fourier_feature_networks}. Here we compare against the alternative of direct pixel-level representations for RGB image generation. 
    (top) The first row shows images generated using implicit representations while the second row shows those of pixel-level ones. The two graphs below correspond to pixel variance (y-axis) plotted against iteration (x-axis) for implicit and pixel-level representation respectively from left to right.}
    \label{fig:ablate_neural}
\end{minipage}
\vspace{-1em}
\end{figure*}
\begin{figure*}[!h]
\centering
\begin{minipage}{\linewidth}
    \includegraphics[width=\linewidth]{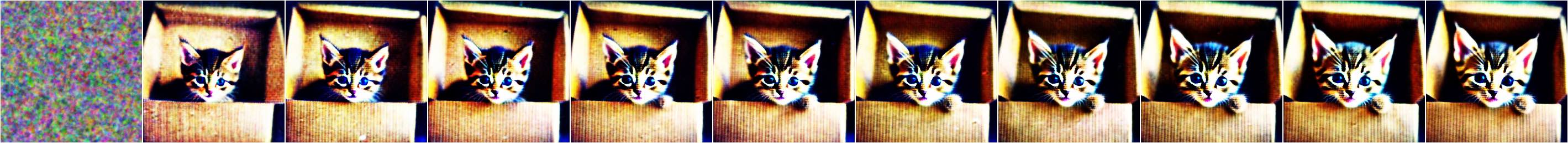}\\
    \includegraphics[width=\linewidth]{figures__vis_appendix__raster_cat.jpg} \\
    \hspace*{\fill}\small{$\overrightarrow{\texttt{Iteration}}$}\hspace*{\fill} \\
\end{minipage}
\begin{minipage}{0.25\linewidth}
    \includegraphics[width=\linewidth]{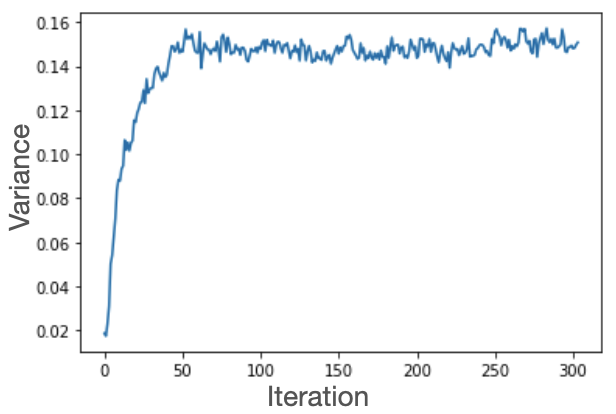}
\end{minipage}
\hspace{0.01\linewidth}
\begin{minipage}{0.72\linewidth}
    \caption{\textbf{Visualizing implicit neural representation}:
    We illustrate another example of generating an image using our implicit neural representation (top) vs pixel-wise representation (bottom). To the left we show a graph of pixel level variance (y-axis) plotted against iteration (x-axis) for the implicit representation case. In this example, we highlight how the characteristics of the cat (shape, location) is consistent in the pixel-wise case from early iterations. However, in our implicit (parametric) representation, various characteristics of the cat (e.g. size, position) alter throughout the iterations. This latter kind of behaviour is also favourable to our task of mask generation.}
    \label{fig:ablate_neural1}
\end{minipage}
\vspace{-1em}
\end{figure*}

In \modelname's \vart{Fourier} and \vart{Bilateral Fourier} variants, as well as our transparent image generations, we use the neural-neural texture formulation from \cite{Burgert2022} to represent our learnable alpha masks (and RGB channels in transparent image generations, and foreground/background in our analogous example in \cref{sec:analagous}). The alternative to using this implicit neural representation is learning the pixels directly by representing them as a learnable tensor of dimensions $(H,W,C)$ for $C=1$ in mask and $C=3$ in RGB images. In this ablation, we explore how the alternative compares against our selected approach. We highlight the two main drawbacks of pixel-level representations as 1) noisy outputs and 2) bad convergence. 

We illustrate this behavior in \cref{fig:ablate_neural}. First, we attempt to generate images using each of the two image parametrizations through score distillation sampling.  We show that neural representations lead to less noisy outputs. The top two rows in \cref{fig:ablate_neural} correspond to these. Clearly, the neural representation in the first row leads to a less noisy output in comparison to the alternative. Next, we explore the convergence behavior of each approach. To analyze this, we utilize the images being generated and measure their variance at the pixel level. While a natural image contains some variance in pixel space, this is a finite value, and our run-time optimization should ideally converge at this variance. However, utilizing a pixel-level representation results in continuously increasing variance in the image pixels, leading to overly saturated images (dissimilar to a realistic image) and in the case of masks, lack of convergence. This is illustrated in the two graphs at the bottom of \cref{fig:ablate_neural}. We also highlight how in these graphs of \cref{fig:ablate_neural} the variance of the implicit representation converges early on at around 50 iterations (left) while that of the pixel-wise representation (right) fails to converge even after 350 iterations.  

\subsection{Bilateral Filter}
\begin{figure}[h]
\centering
\begin{minipage}{\linewidth}
    \includegraphics[width=\linewidth]{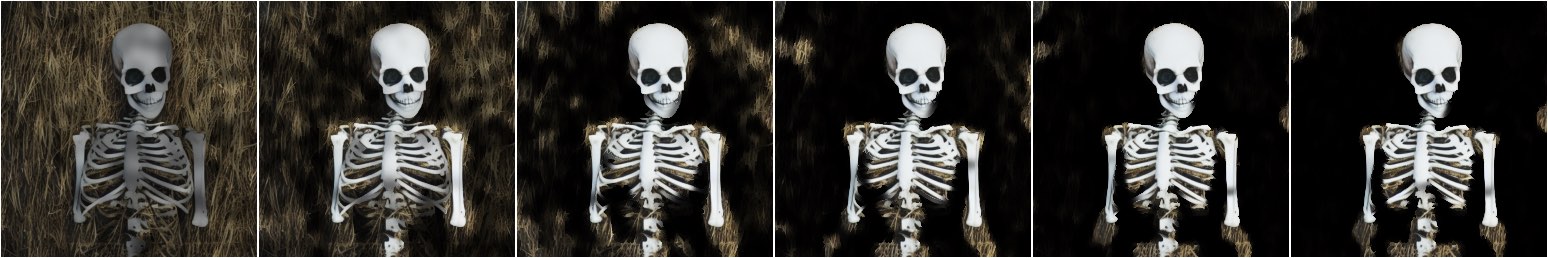}\\
    \includegraphics[width=\linewidth]{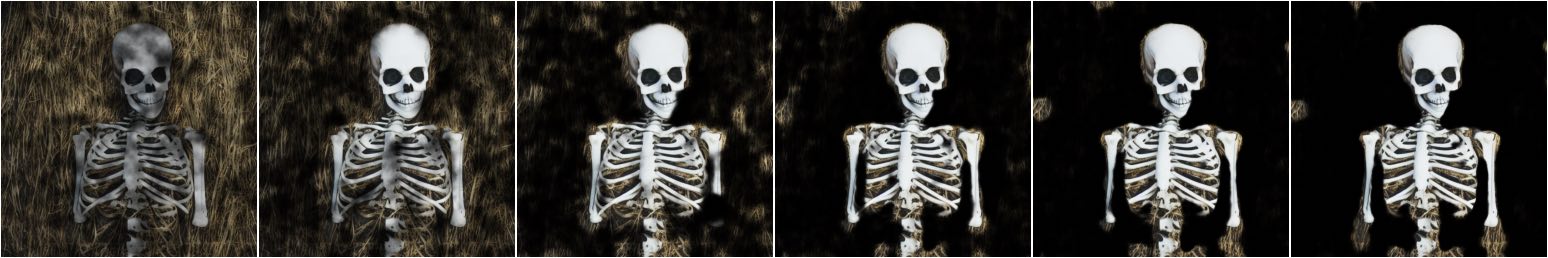} \\
    \includegraphics[width=\linewidth]{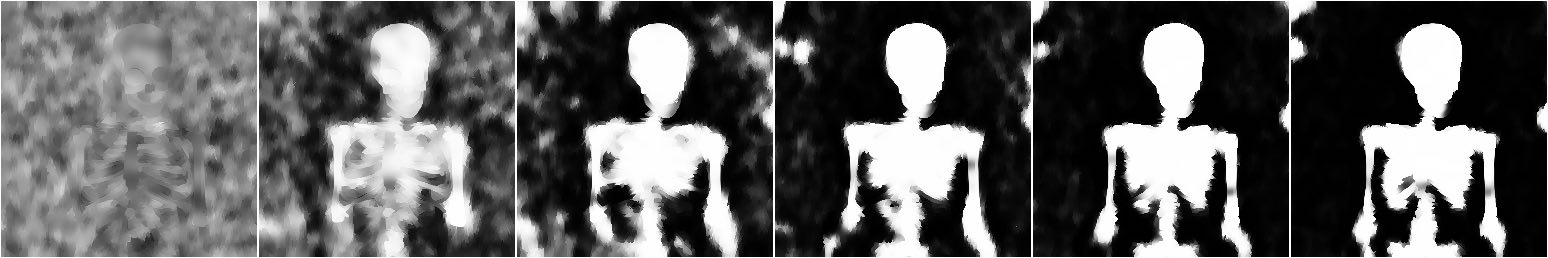}\\
    \includegraphics[width=\linewidth]{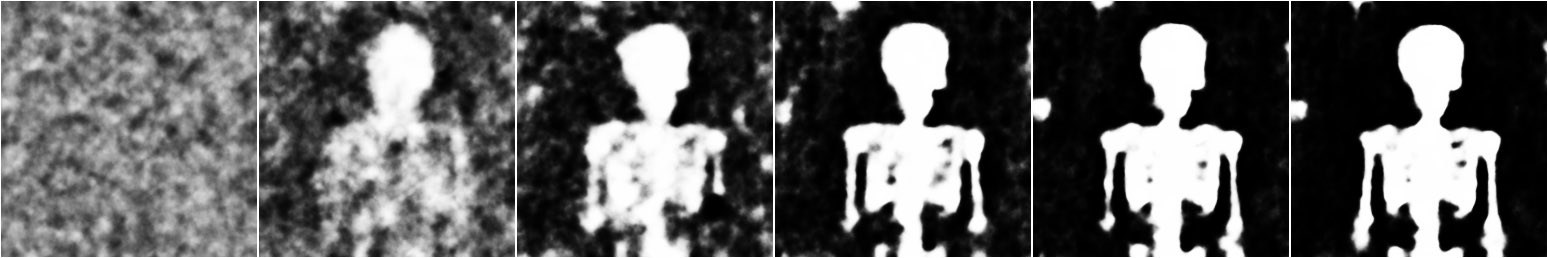}\\
    \hspace*{\fill}\small{$\overrightarrow{\texttt{Iteration}}$}\hspace*{\fill} \\
\end{minipage}
\begin{minipage}{\linewidth}
    \caption{\textbf{Ablation on bilateral filters}:
    The bilateral filter improves segmentation alignment with object boundaries, resulting in more accurate and precise segmentations. In this figure, we show a timelapse of the alpha mask optimization process over time from left to right, for both Peekaboo variants \vart{Fourier} and \vart{Bilateral Fourier} (see \cref{sec:experiments}). The bottom two rows show a timelapse of the alpha maps, and the top two rows show a timelapse of those alpha maps overlaid on the original image to help visualize their accuracy. The first and third rows depict the \vart{Fourier} variant, while the second and fourth rows depict the \vart{Bilateral Fourier} variant.
    }
    \label{fig:ablate_bilateral}
\end{minipage}
\vspace{-1em}
\end{figure}
In \cref{sec:bilateralfiltersubsub}, we discussed Peekaboo's use of a bilateral filter as part of the image parametrization. In this section, we provide a visualization of this filter.
A closer look at the noisy mask in row 3 column 1 of \cref{fig:ablate_bilateral} will show vague outlines of a skeleton within the noise; this is a result of the modified bilateral filter being applied to the mask.

\section{Optimization Details}
In this section, we discuss all details relevant to our proposed inference time optimization that generates segmentations. In order to generate a single segmentation, we run it for 200 iterations, a learning rate of $1e-5$, and stochastic gradient descent as the optimizer.

Additionally, we apply the modified bilateral blur operation, conditioned on the image to be segmented, onto the learnable alpha masks at initialization, which results in faster and better convergence. Here we use a blur kernel of size 3 with 40 iterations (multiple iterations increase the effective field of view for a kernel).

\section{Limitations}  
\begin{figure*}[t]

\begin{minipage}{\linewidth}
	\includegraphics[width=0.19\linewidth]{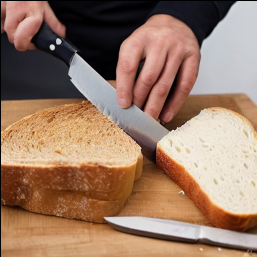}
	\includegraphics[width=0.19\linewidth]{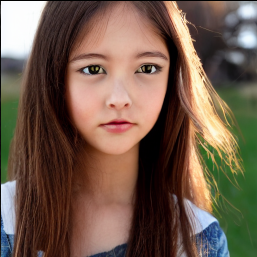}
	\includegraphics[width=0.19\linewidth]{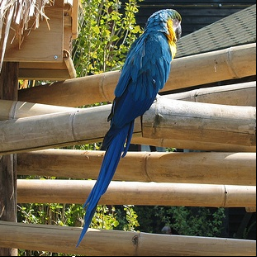}
	\includegraphics[width=0.19\linewidth]{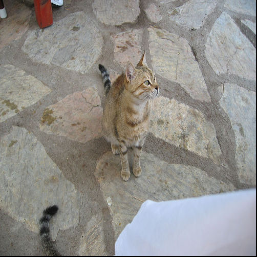}
	\includegraphics[width=0.19\linewidth]{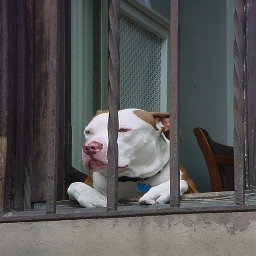}
\end{minipage}
\begin{minipage}{\linewidth}
	\includegraphics[width=0.19\linewidth]{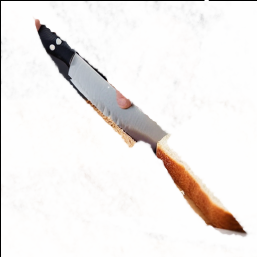}
	\includegraphics[width=0.19\linewidth]{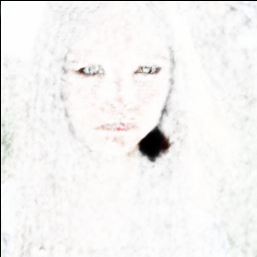}
	\includegraphics[width=0.19\linewidth]{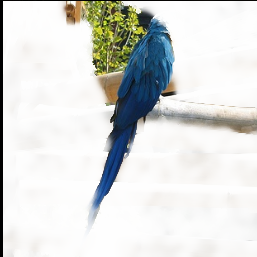}
	\includegraphics[width=0.19\linewidth]{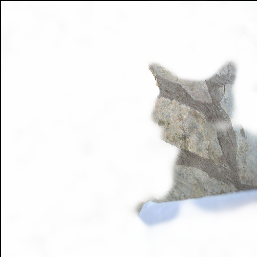}
	\includegraphics[width=0.19\linewidth]{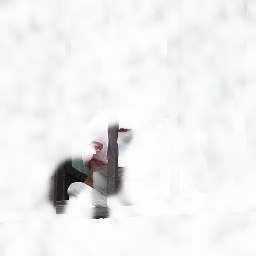}
\end{minipage}
\caption{\textbf{Limitations and failures of \modelname }:
We illustrate examples where \modelname fails to properly segment the region of interest. The prompts for these examples from left to right are: \texttt{knife}, \texttt{eyes}, \texttt{bird}, \texttt{cat}, \texttt{dog}. We note how a major issue is the tendency of \modelname to hallucinate the shape of the object denoted by the text caption.  
}
\label{fig:vis_lim}

\end{figure*}

We also acknowledge that our proposed \modelname contains various limitations and shortcomings. A key drawback is its failure cases that result in a hallucination of the text prompt using some random background region, i.e. it uses the background texture to create a region shaped like the underlying object described by the text. We illustrate this behavior in \cref{fig:vis_lim}. This is clearly visible in the three right-most examples (bird, cat, dog). We also note that such behavior is more common when a simple (often one word) text caption is used. Another failure is the addition of unnecessary parts to the region of interest. For example, in column one of \cref{fig:vis_lim}, while the knife is coarsely localized, \modelname also incorrectly creates a handle for it using a slice of bread from the background. The model also sometimes fails to converge entirely, as illustrated in the second column, where despite localizing the eyes, the generated mask also holds onto outlines of the image foreground object. 

While we hope to address these issues in future work, we also reiterate that despite these limitations, \modelname is a first unsupervised method that is able to perform open vocabulary segmentation using arbitrary natural language prompts.

\newpage

\section{Analagous Example}
\label{sec:analagous}
\begin{figure}[!h]
\centering
\includegraphics[width=0.75\linewidth]{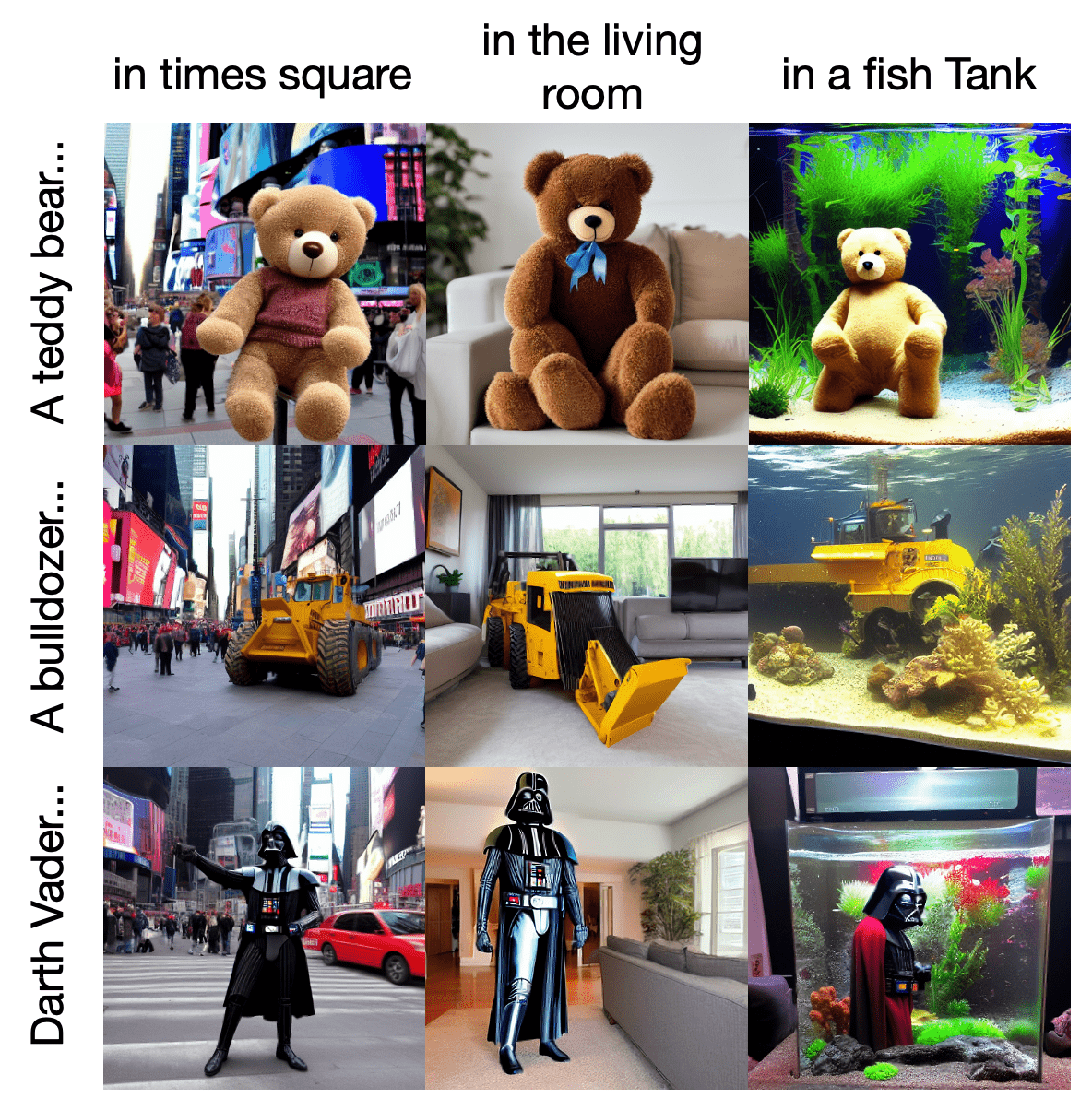}
\caption{\textbf{Images generated from Stable Diffusion:} 
This data distribution is quite different from the distribution of natural images, rarely containing objects in non-central locations of the image. This bias is carried through to Peekaboo, which is why Peekaboo is best at segmenting objects near the center of an image. These images were generated by running prompts through stable diffusion v1.4, such as ``a teddy bear in times square'', and using DDPM with a guidance scale of 7. }
\label{fig:minigrid}
\end{figure}
\begin{figure*}[!h]
\centering
\begin{minipage}{0.96\textwidth}
\vspace{-1.0em}
\includegraphics[width=0.95\textwidth]{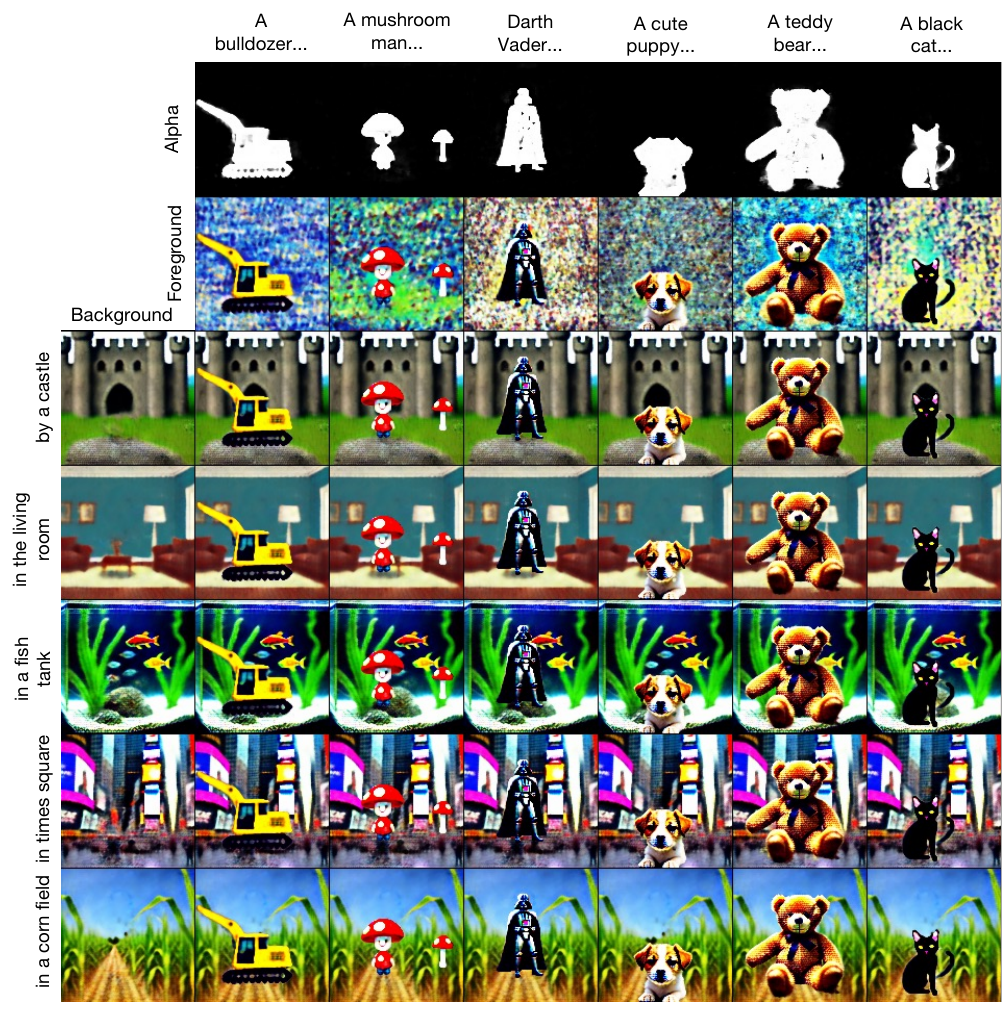}
\end{minipage}
\begin{minipage}{0.96\textwidth}
\caption{\textbf{Can diffusion models separate foreground and background?} 
Stable diffusion was only trained on RGB images. However, observing the often clean boundaries between objects in generated images, it begs the question: can we use these boundaries to generate images with transparency?
In this figure, we conduct a self-contained experiment that answers this question: yes.
We overlay 6 learnable foreground images on top of 5 learnable background images using 6 learnable alpha masks, to get a total of 30 learnable composite images. Each of these composite images has a composite caption, created by combining the foreground prompt with the background prompt. For example, the image created by combining background \#1 and foreground \#1 is accompanied by the prompt \texttt{"a bulldozer by a castle"}.
We optimize each of these 30 composite cells with score distillation loss (see \cref{subsubsec:sds}) until each foreground, background and alpha mask has been learned.
Differences between this experiment and \modelname are minimal: in Peekaboo, we only optimize the alpha masks, whereas in this analogous experiment we optimize everything.
%
%
}
\label{fig:motivation}
\end{minipage}
\vspace{-0.5em}
\end{figure*}

\begin{figure*}[!h]
\centering

\includegraphics[width=\textwidth]{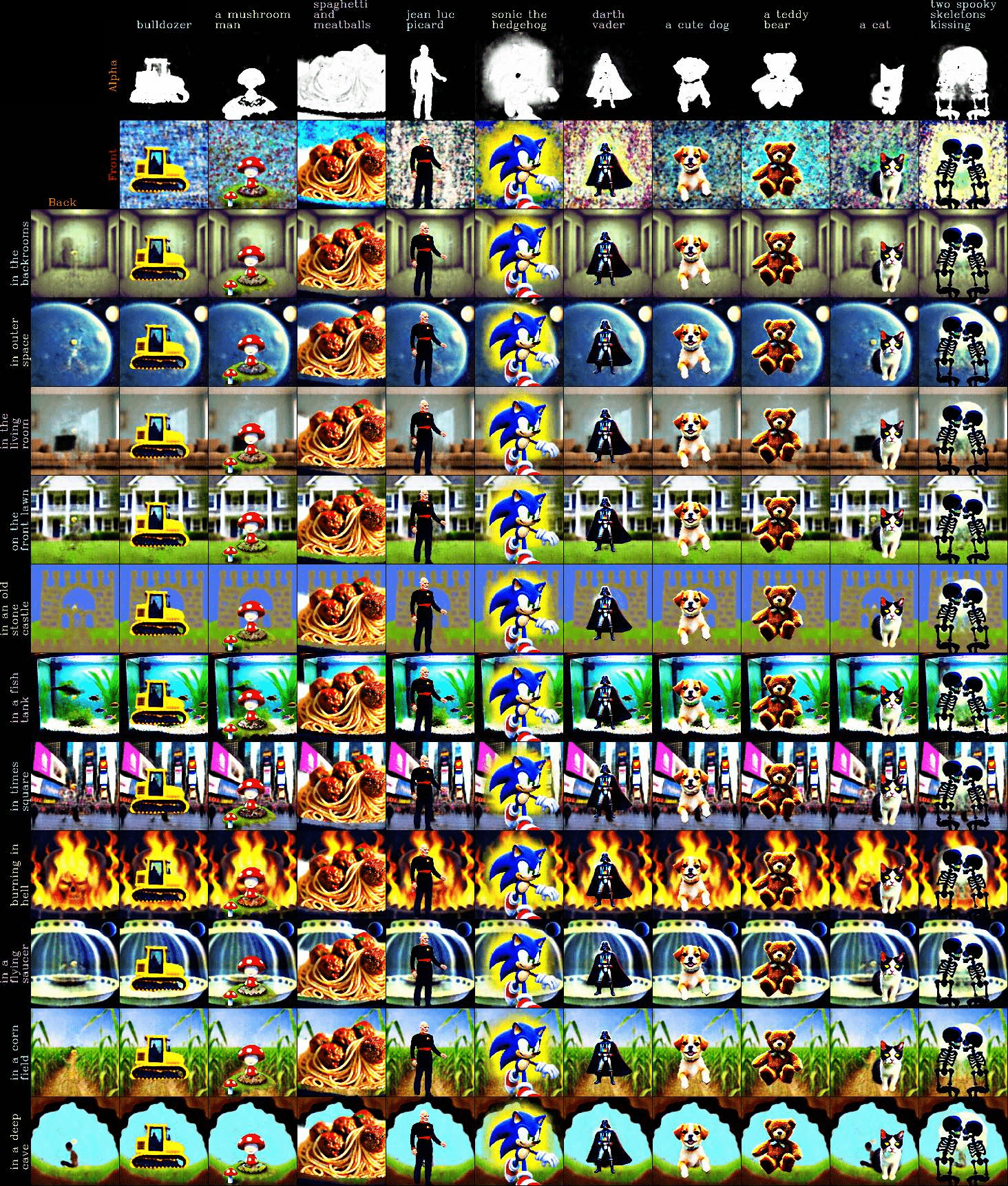}
\caption{\textbf{Extended visualization of analogous experiment}
}
\label{fig:big_grid}
\end{figure*}
In order to explain the intuition behind Peekaboo, we show results of analogous experiment that helped to inspire it. 
This analagous algorithm is not exactly Peekaboo, but it is very similar and is helps explain the main idea behind how Peekaboo works.

We first examine a stable diffusion model \cite{Rombach2022HighResolutionIS} pre-trained on LAION-5B \cite{laion5b}. Our goal is to explore whether internal knowledge of these models regarding boundaries and localization of individual objects can be accessed and subsequently utilized for tasks such as segmentation.
We focus on the case of generating a single synthetic object in some background and attempt to generate an accompanying alpha mask that demarcates the region belonging to the foreground object. We utilize score distillation loss (see \cref{subsubsec:sds}) as a cross-modal similarity function that connects a text caption describing a foreground object to the image region it is located. Using this similarity as an optimization objective, we generate the foreground, background, and alpha mask. 

Results obtained from this process are illustrated in \cref{fig:motivation}. While our method is able to generate good segmentation masks relevant to the foreground, we note that generated images are unrealistic. 

We highlight that generated image quality is indifferent to the segmentation component, where we generate single-channel alpha masks. The rest of our work is focused on how this technique can be leveraged for segmenting stand-alone images, i.e. images beyond those generated by a diffusion model. In essence, we attempt to segment real-world images with free-form text captions.

\newpage

\begin{figure*}[H]
\centering

\includegraphics[width=.9\textwidth]{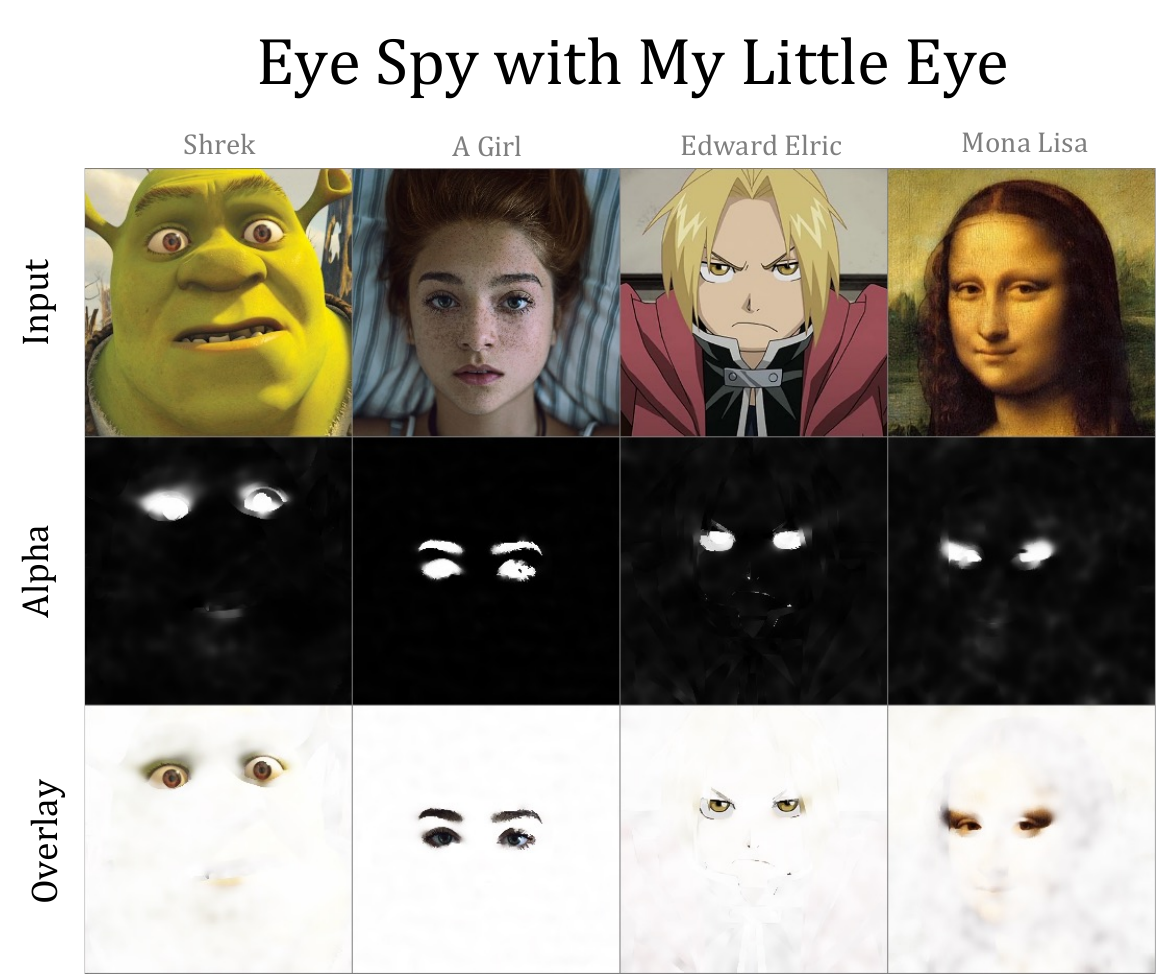}
\caption{In this figure, we play a game: find the eyes! Peekaboo can segment specific features of images, and is good at finding eyes. Only one prompt was used: \texttt{eyes}. On the top row we have input images, and in the middle row we have the outputted alpha map. On the bottom we overlay the input images with a white mask corresponding to the alpha, to better show which part of the image it chose to segment.
}
\label{fig:eye_spy}
\end{figure*}

\section{More Results}

\begin{figure*}[!h]
\centering

\includegraphics[width=\textwidth]{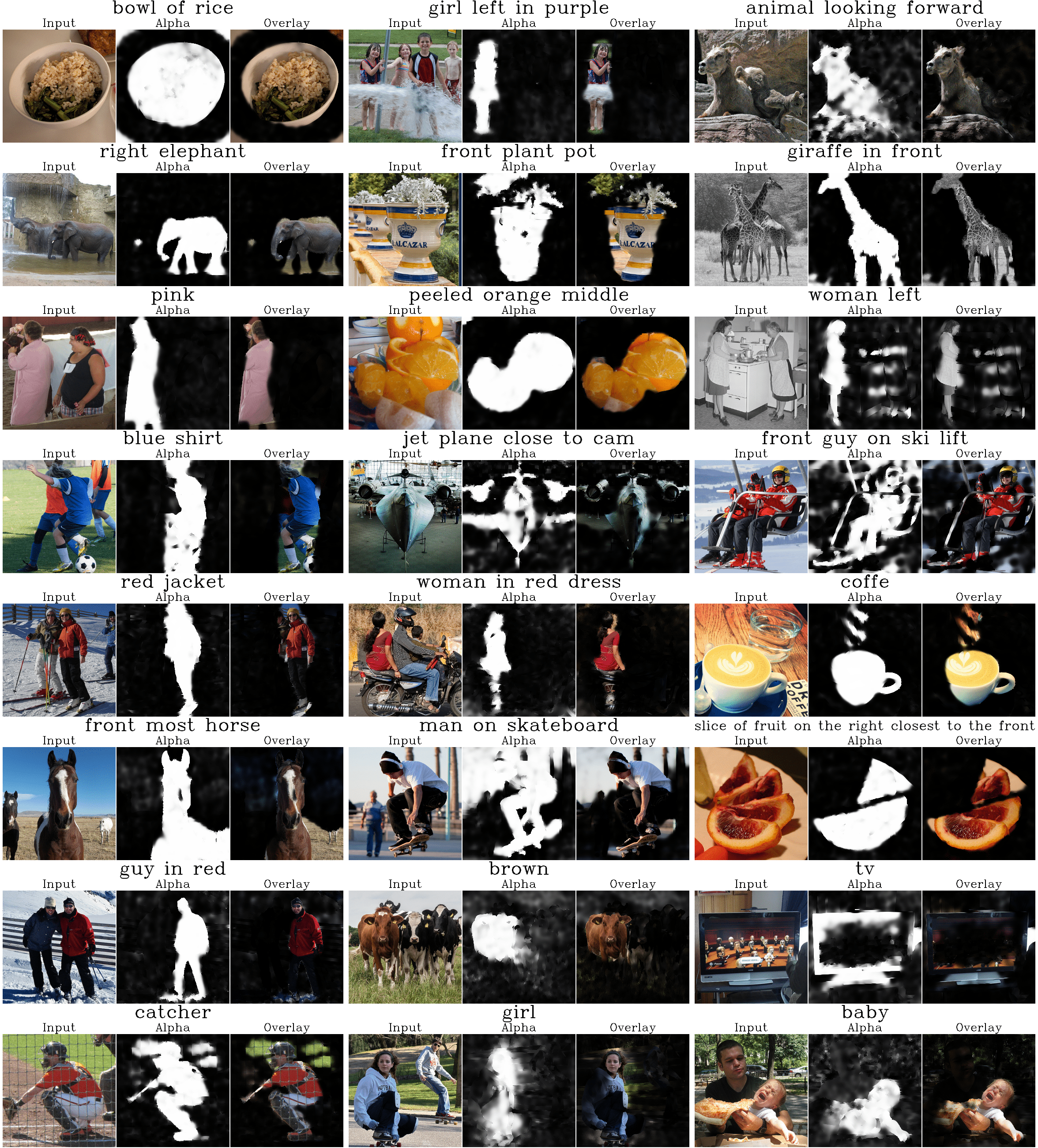}
\caption{Our results on RefCOCO-C. For each sample we demonstrate the prompt text(title), input image(left), our output alpha mask(middle), and the image segmented by the mask(right)
}
\label{fig:coco_res}
\end{figure*}
\begin{figure*}[!h]
\centering

\includegraphics[width=0.45\textwidth]{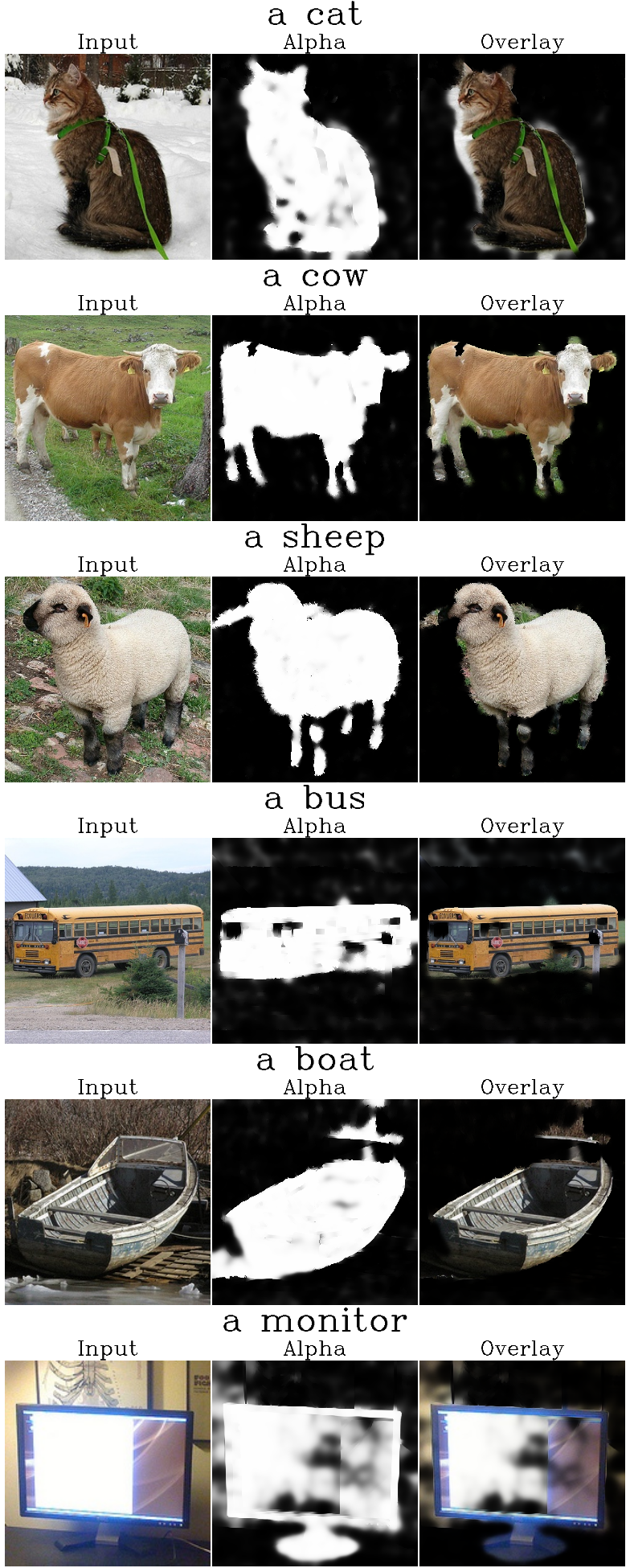}
\caption{Our results on Pascal VOC-C. For each sample we demonstrate the prompt text(title), input image(left), our output alpha mask(middle), and the image segmented by the mask(right)
}
\label{fig:voc_res}
\end{figure*}
\begin{figure*}[!h]
    \centering
    
    \includegraphics[width=.6\textwidth]{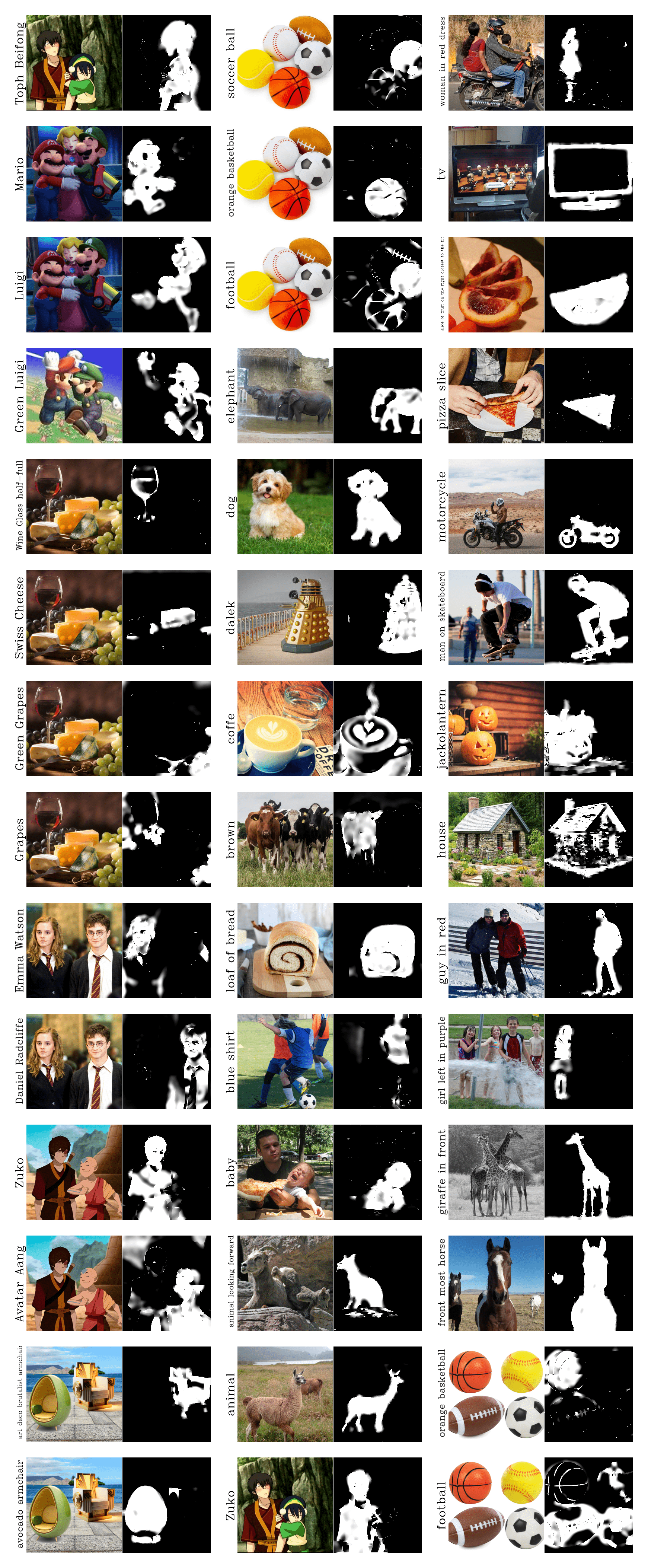}
    \caption{Peekaboo's segmentation results on various images, including pop references such as Avatar the Last Airbender and AI-generated images of imaginary objects that don't exist such as avocado armchairs. Prompt and input image are on the left, and the alpha mask output is on the right.
    }
    \label{fig:coco_res}
\end{figure*}


\newpage
    \begin{figure*}[!h]
        \centering
        
        \includegraphics[width=.96\textwidth]{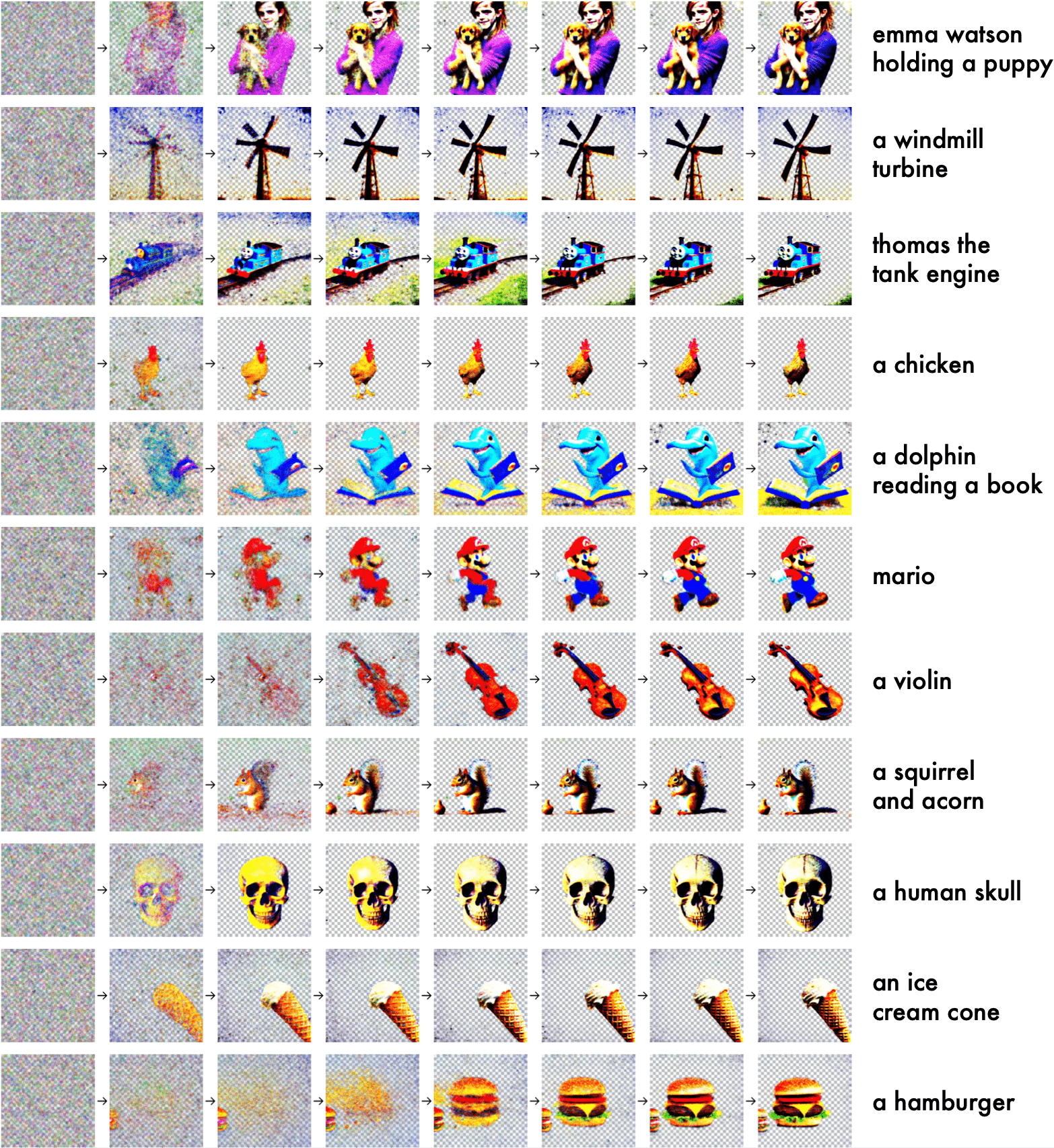}
        \caption{\textbf{Part 1/2.} Peekaboo can generate images with transparency masks! Continuing from \cref{fig:rgbaexamples}, we display timelapses of the transparent image generation task described in \cref{sec:rgbagen}. The prompt for each image is to its right.
     }
        \label{fig:extrargbapart1}
    \end{figure*}
    
    \newpage

    \begin{figure*}[!h]
        \centering
        
        \includegraphics[width=.96\textwidth]{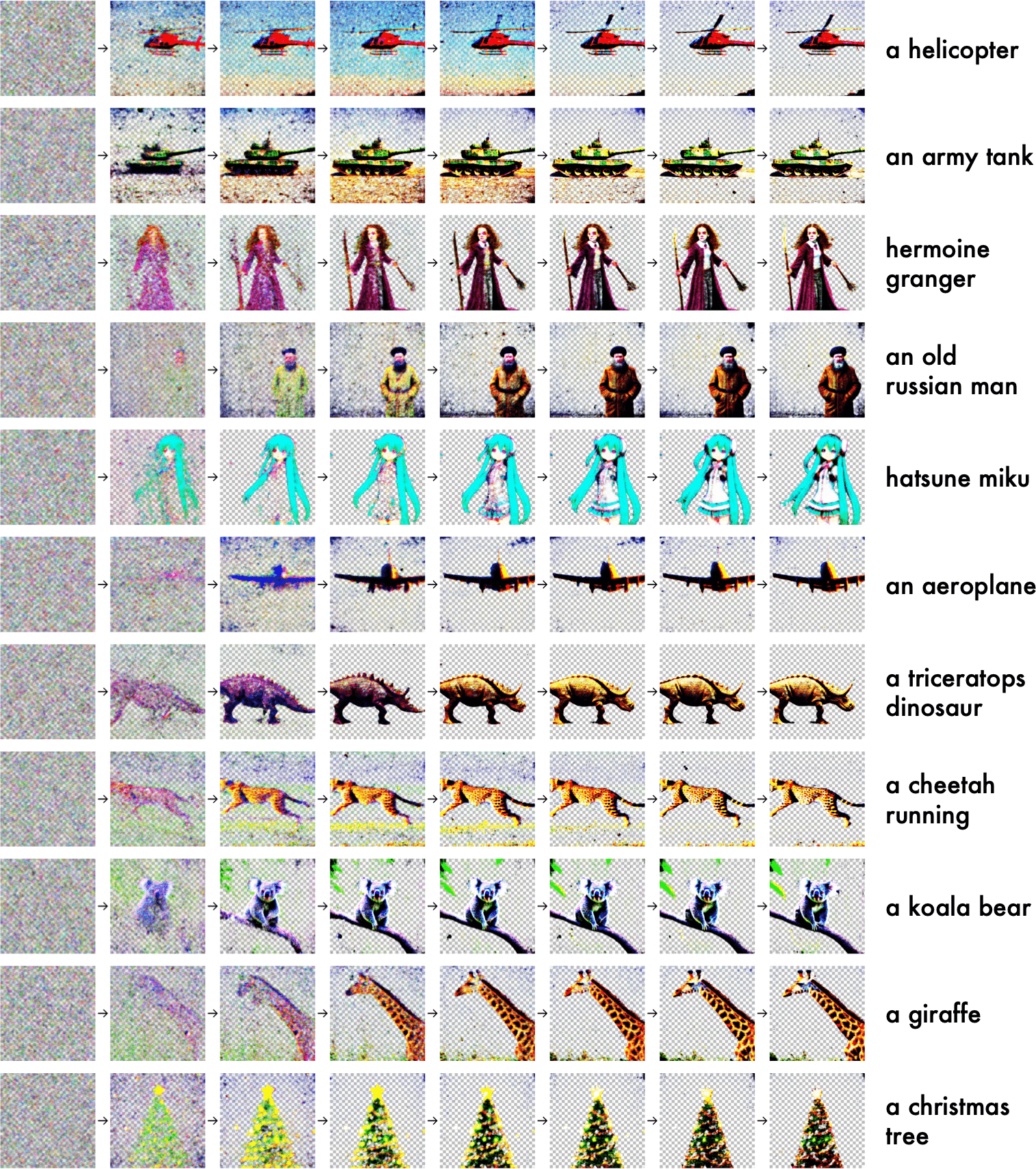}
        \caption{\textbf{Part 2/2.} Peekaboo can generate images with transparency masks! Continuing from \cref{fig:rgbaexamples}, we display timelapses of the transparent image generation task described in \cref{sec:rgbagen}. The prompt for each image is to its right.
         }
        \label{fig:extrargbapart1}
    \end{figure*}

\end{document}